\documentclass{article}

\usepackage{microtype}
\usepackage{graphicx}
\usepackage{subfigure}
\usepackage{amsfonts}       
\usepackage{nicefrac}       
\usepackage{floatrow}
\usepackage{xcolor}
\usepackage{amsmath, amsthm, amssymb, centernot}
\usepackage{booktabs, rotating, multirow} 
\usepackage{sidecap}
\newtheorem{define}{Definition}

\newtheorem{prop}{Proposition}

\newtheorem{remark}{Remark}
\newtheorem{coroll}{Corollary}
\DeclareFloatFont{tiny}{\small}
\newcommand{\histogram}{
\begin{figure}[t!]
\includegraphics[scale=0.25]{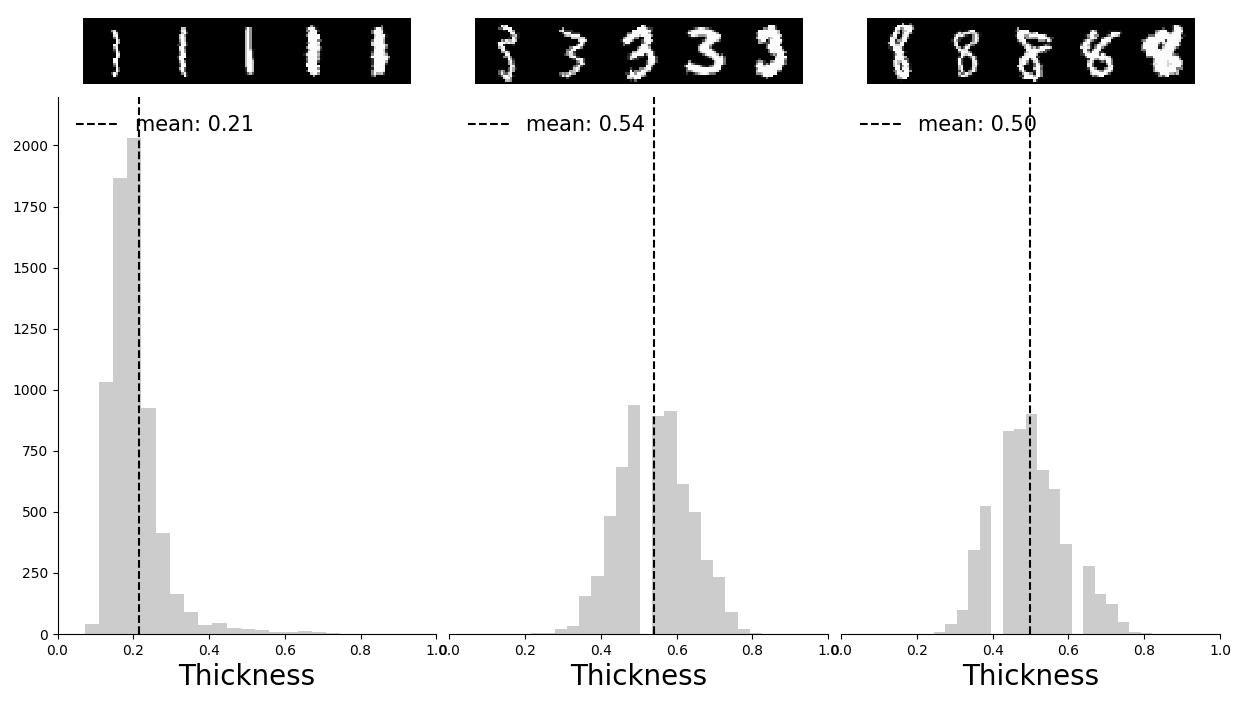}
\vspace{-0.2in}
\caption{Empirical distributions for the continuous state variable representing stroke width are digit-dependent, illustrating dependence of style (width) on type (digit).}
\label{fig:histogram}
\end{figure}
}

\newcommand{\introFigure}{
\begin{figure}[t!]
\includegraphics[width=\columnwidth]{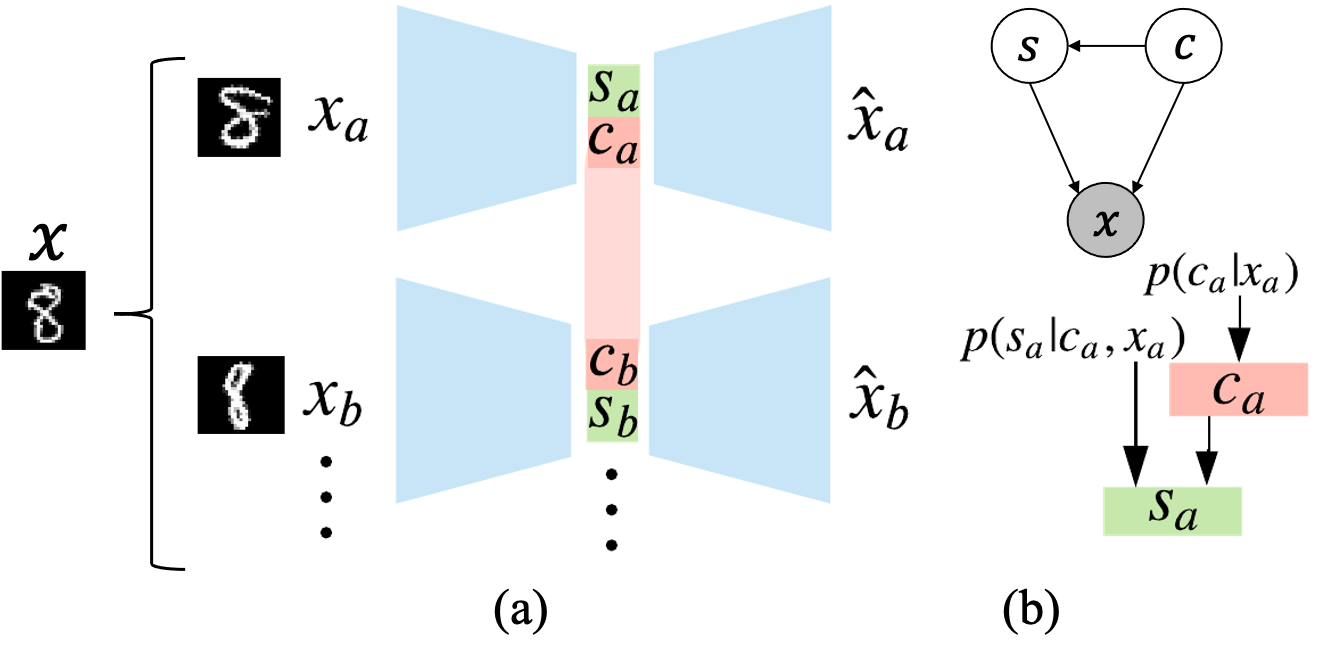}
\caption{(a) Multi-arm autoencoder framework proposed as cpl-mixVAE model. Individual arms receive non-identical noisy copies of given samples $\mathbf{x}$, i.e. $\{\mathbf{x}_a, \mathbf{x}_b$, $\dots\}$, where they all belong to the same category, to learn mixture representations, i.e. $\{q(\mathbf{c}_a, \mathbf{s}_a)$, $q(\mathbf{c}_b,\mathbf{s}_b)$, $\dots\}$. VAE arms cooperate to learn the categorical assignment, $p(\mathbf{c})$. Cooperation is achieved by imposing a penalty on mismatches in the categorical assignments. (b) Each autoencoder learns type dependence of the state variable according to the graphical model.}
\vspace{-0.3in}
\label{fig:introFigure}
\end{figure}
}

\newcommand{\allDataLatentTraversa}{
\begin{figure*}[t!]
\centering
\includegraphics[scale=0.4,trim=4 4 4 4,clip]{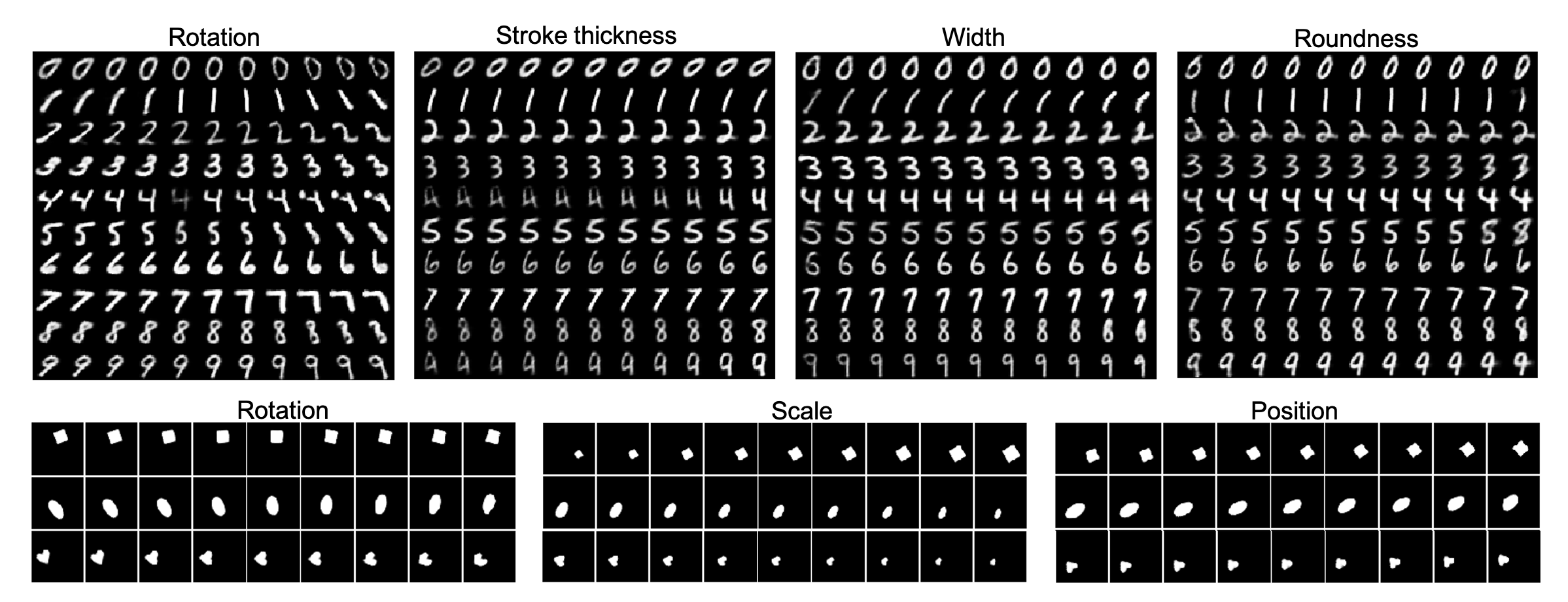}
\vspace{-0.2in}
\caption{Interpretable continuous latent traversals of 1-st arm of the cpl-mixVAE framework with two autoencoders, for MNIST (top panel) and dSprites (bottom panel). The discrete variable $\mathbf{c}$ is constant for all reconstructions in the same row.}
\label{fig:allDataLatentTraversa}
\end{figure*}
}

\newcommand{\clusteringFACSall}{
\begin{figure*}[ht]
\includegraphics[scale=0.5, trim=6 6 6 6, clip]{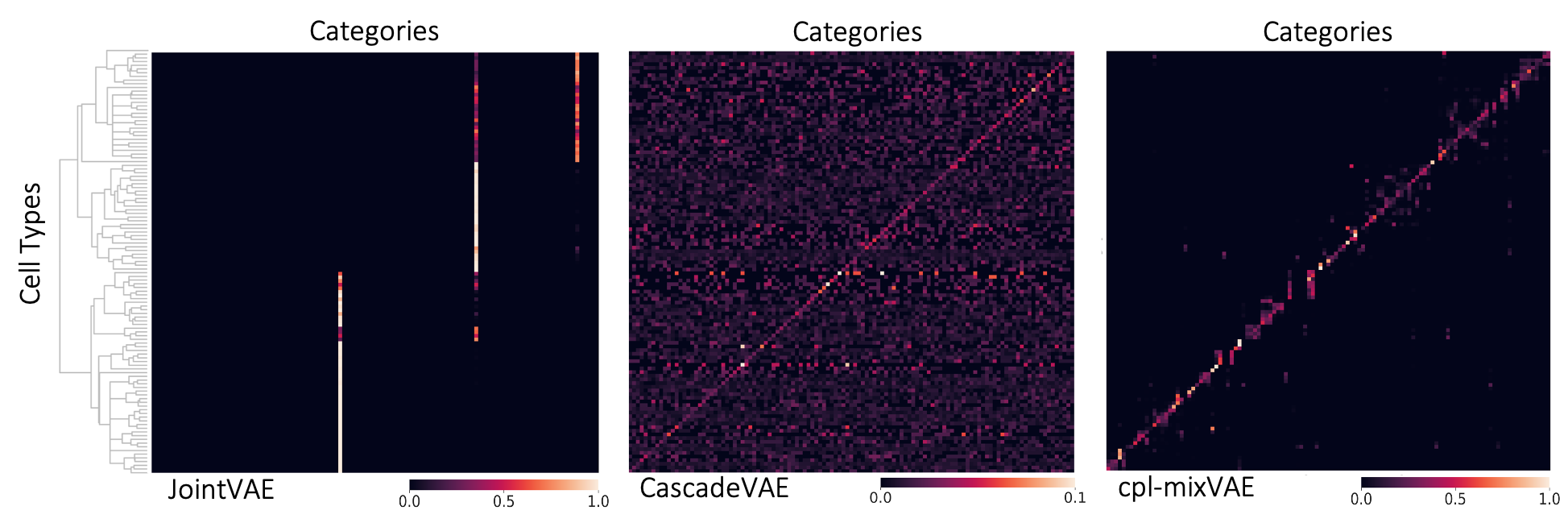}
\vspace{-.2in}
\caption{Categorical assignments for the scRNA-seq dataset. Confusion matrices of JointVAE, CascadeVAE, and cpl-mixVAE trained by $|\mathbf{c}|=115$, $|\mathbf{s}|=2$. The dendrogram on the y-axis shows MG-based hierarchical classification with $115$ cell types, suggested by Tasic et al., 2018. Implementation details for each model can be found in Supplementary Section J.}
\label{fig:clusteringFACSall}
\end{figure*}
}

\newcommand{\latentTraversalFACS}{
\begin{figure}[b!]
\begin{center}
\centering
\vspace{-0.1in}
{\includegraphics[width=\columnwidth,trim=6 6 6 6,clip]{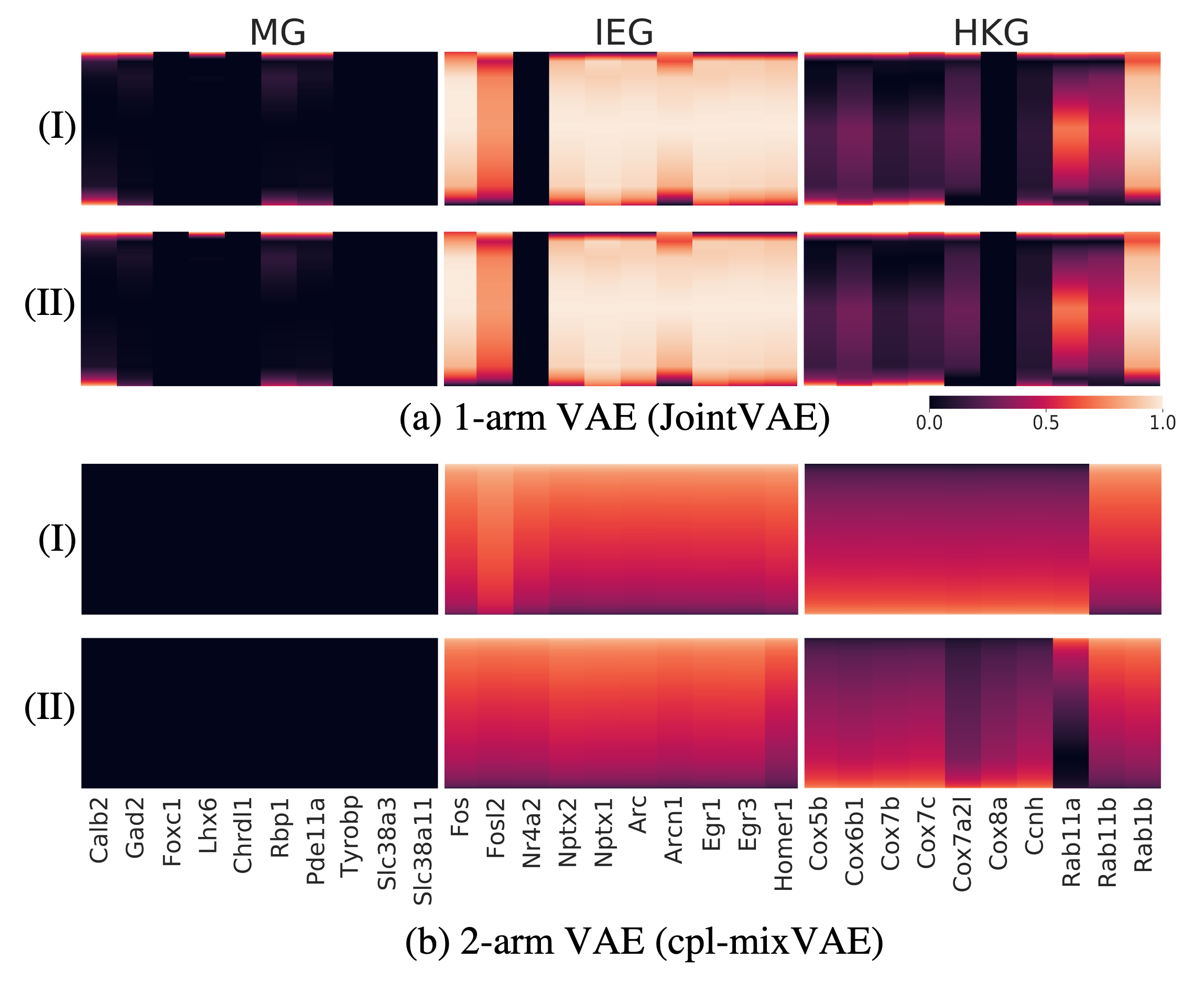}}
\vspace{-.3in}
\caption{Continuous latent traversal analysis for two excitatory cell types (I) ``L5 NP ALM Trhr Nefl'' and (II) ``L6 CT Nxph2 Sla'', for 1-arm VAE (JointVAE) and 2-arm VAE (cpl-mixVAE). For each type, the traversal is color-mapped to a normalized reconstructed gene expression value (colorbar) as a function of the state variable for 3 gene subsets: marker genes (MG), immediate early genes (IEG), and housekeeping genes (HKG).}
\label{fig:latentTraversalFACS}
\end{center}
\vspace{-.25in}
\end{figure}
}

\newcommand{\multiagentFACS}{
\begin{figure}[b!]
\vspace{-0.1in}
\floatbox[{\capbeside\thisfloatsetup{capbesideposition={right,top},capbesidewidth=3.5cm}}]{figure}
{\caption{Improvement of the categorical representation (ACC) of cpl-mixVAE by adding more agents to the multi-arm framework. A-arm’s performance for $A \ge 2$ is compared with the baseline 1-arm, JointVAE. The reported results belong to 3 randomly initialized runs. }\label{fig:multiagentFACS}}
{\includegraphics[scale=0.42]{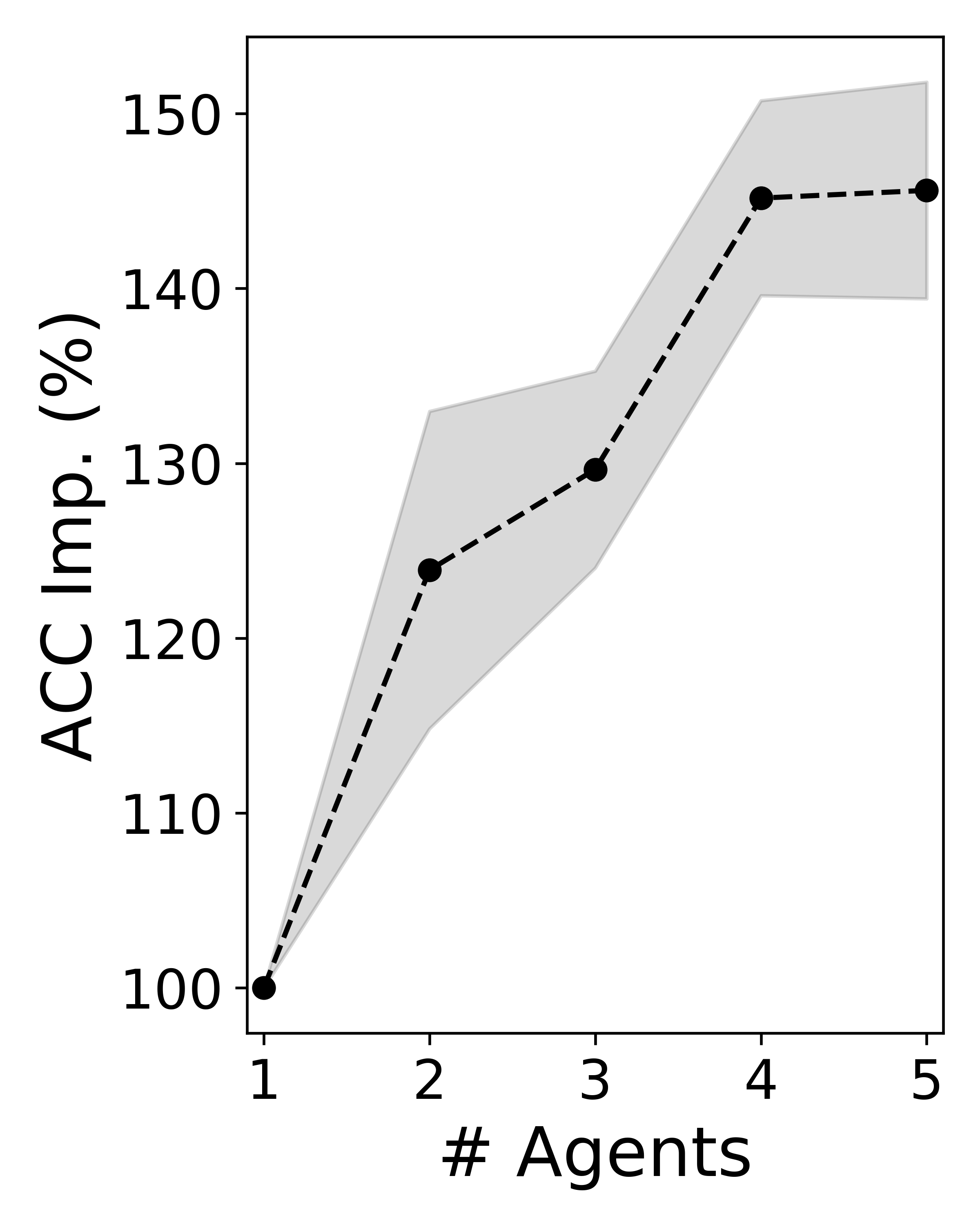}}
\vspace{-0.2in}
\end{figure}
}

\newcommand{\allResults}{
{\small{\begin{table*}[t!]
\centering
\vspace{-0.05in}
\caption{Training results for 10 randomly initialized runs. cpl-mixVAE uses 2 arms. $|\mathbf{c}|$ and $|\mathbf{s}|$ denote the cardinality of latent discrete and continuous spaces. Chance-level indicates the chance level of classification accuracy for each dataset. ACC denotes the accuracy of the categorical assignment. Computation denotes the training speed~(iteration/second) on a GeForce RTX 2080 Ti GPU. The computation of cpl-mixVAE includes the entire execution time for training one pair of coupled networks, plus data augmentation.}
\label{tab:allResults}
\begin{tabular}{|l c c |c l l c| c|} 
\hline 
 \raisebox{-1.2ex}{\bf{Dataset}} & \raisebox{-1.2ex}{\bf{Chance-level}} & \raisebox{-1.2ex}{$|\mathbf{c}|$} & \raisebox{-1.2ex}{$|\mathbf{s}|$}  & \raisebox{-1.2ex}{\bf{Method}}  &\multicolumn{1}{c}{\raisebox{-.3ex}{{\bf{ACC}~($\%$)}}~$\uparrow$} & \raisebox{-.3ex}{{\bf{Computation}}~$\uparrow$} & \raisebox{-.3ex}{\bf{Disentanglement}} \\[-1ex]
 &  &   & & & {(\text{mean} $\pm$ s.d.)} & (iteration/sec) & {\bf{score}}\\
\hline 
& & &  \raisebox{-.2ex}{2} & \raisebox{-.2ex}{InfoGAN}  &  \raisebox{-.2ex}{77.87 $\pm$ 21.68} & \raisebox{-.2ex}{12.2}  &\\\cline{4-7}
& &  & & \raisebox{-.5ex}{JointVAE}  &  \raisebox{-.5ex}{68.99 $\pm$ 11.76} & \raisebox{-.5ex}{74.1}  &\\[-0.5ex] 
\raisebox{1.3ex}{MNIST} & \raisebox{1.3ex}{10.0$\%$} & \raisebox{1.3ex}{10} &  10  & CascadeVAE  &  81.41 $\pm$ 09.54 & 23.8 &  \raisebox{1.3ex}{--}\\ 
& &  & & cpl-mixVAE  &  \bf{84.56 $\pm$ 06.47} &  17.5 &\\
\hline 
& &  & & {JointVAE} &  44.79 $\pm$ 03.88 & 52.6 & 74.51 $\pm$ 05.17\\
dSprite & 33.3$\%$ & 3 & 6 & CascadeVAE  &  78.84 $\pm$ 15.65 & 15.4 & {90.49 $\pm$ 05.28}\\ 
& &  & & cpl-mixVAE  & \bf{96.30 $\pm$ 09.15} & 20.6 & 89.98 $\pm$ 04.09\\
\hline
& &  & & JointVAE & 12.53 $\pm$ 01.83 & 28.6 &\\
scRNA-seq & 06.3$\%$ & 115 & 2  & CascadeVAE  &  02.69 $\pm$ 00.05 & 03.4 & --\\ 
& &  & & cpl-mixVAE &  \bf{38.78 $\pm$ 01.26} & 10.1 &\\
\hline
\end{tabular}
\vspace{-0.1in}
\end{table*}}}
}

\def\arrvline{\hfil\kern\arraycolsep\vline\kern-\arraycolsep\hfilneg}

\usepackage{hyperref}

\usepackage[accepted]{icml2021}

\icmltitlerunning{Mixture Representation Learning with Coupled Autoencoders}

\begin{document}

\twocolumn[
\icmltitle{Mixture Representation Learning with Coupled Autoencoders}




\begin{icmlauthorlist}
\icmlauthor{Yeganeh M. Marghi}{to}
\icmlauthor{Rohan Gala}{to}
\icmlauthor{Uygr S\"{u}mb\"{u}l}{to}
\end{icmlauthorlist}

\icmlaffiliation{to}{Allen Institute, WA, USA}

\icmlcorrespondingauthor{Yeganeh M. Marghi}{yeganeh.marghi@alleninstitute.org}

\icmlkeywords{Machine Learning, ICML}

\vskip 0.3in
]



\printAffiliationsAndNotice{} 

\begin{abstract}
Jointly identifying a mixture of discrete and continuous factors of variability without supervision is a key problem in unraveling complex phenomena. Variational inference has emerged as a promising method to learn interpretable mixture representations. However, posterior approximation in high-dimensional latent spaces, particularly for discrete factors remains challenging. Here, we propose an unsupervised variational framework using multiple interacting networks called \textit{cpl-mixVAE} that scales well to high-dimensional discrete settings. In this framework, the mixture representation of each network is regularized by imposing a consensus constraint on the discrete factor. We justify the use of this framework by providing both theoretical and experimental results. Finally, we use the proposed method to jointly uncover discrete and continuous factors of variability describing gene expression in a single-cell transcriptomic dataset profiling more than a hundred cortical neuron types.
\end{abstract}
\section{Introduction}
\label{sec:Introduction}
Complex phenomena can be attributed to a mixture of discrete and continuous factors of variability. It is crucial to parse complexity in this manner for a variety of applications, from learning interpretable models for image generation, to quantifying factors of biological variability in single-cell studies. To this end, mixture modeling approaches propose to learn representations that jointly capture dependence of observations on discrete and continuous factors~\cite{bengio2013representation}.

Generative models for mixture modeling have recently received attention from the deep learning community. Deep Gaussian mixture models~\cite{dilokthanakul2016deep, johnson2016composing, jiang2017variational} are among the first deep generative models for mixture modeling, in which a continuous representation is decomposed into discrete clusters. However, such models focus on clustering without regard to interpretability. Various adversarial and variational methods have been proposed to learn interpretable continuous factors alongside: while existing adversarial generative models, e.g. InfoGAN~\cite{chen2016infogan}, are susceptible to stability issues~\cite{higgins2017beta, kim2018disentangling}, variational autoencoders (VAEs) emerge as efficient and more stable alternatives~\cite{tschannen2018recent, zhang2018advances}. 

VAE-based approaches approximate the mixture model by assuming a family of distributions~$q_{\phi}$ and select the member closest to the true model~$p$. Popular choices in VAE implementations include (1) using KL divergence to compute discrepancy between $q_{\phi}$ and $p$, and (2) using a multivariate Gaussian mixture distribution with uniformly distributed discrete and isotropic Gaussian distributed continuous priors. However, such choices may lead to underestimating the posterior variance~\cite{minka2005divergence,blei2017variational}. Solutions to resolve this issue are mainly applicable in low-dimensional spaces or for continuous factors alone~\cite{deasy2020constraining,kingma2016improving,ranganath2016hierarchical,quiroz2018gaussian}.
Yet, the dimension of the latent space can be much larger in many application domains (e.g., cell biology, robotic systems, finance), and learning interpretable mixture representations remains challenging in practice, especially as model complexity increases. For instance, in the explosive field of single cell transcriptomics, hundreds of cell types are implicated as discrete variational factors of thousands of gene expression measurements.

Inspired by \textit{collective decision making}, here we introduce a variational framework using multiple \textit{autoencoding arms} to jointly infer interpretable finite discrete (categorical) and continuous factors in the presence of high-dimensional discrete space. Coupled-autoencoders have been previously studied in the context of multi-modal recordings, where each arm learns only a continuous latent representation for one of the data modalities~\cite{feng2014cross, gala2019coupled, lee2020private}. However, it is not clear whether such architectures can be useful to study standalone (single modality) datasets: (i) Do they provide a fundamental advantage over a single arm? (ii) Does exploration via alignment across arms extend to discrete settings? We answer both of these questions in the affirmative by using pairwise-coupled autoencoders for a single data modality that imposes a consensus constraint on the posterior, in an unsupervised fashion. We demonstrate that by acknowledging the dependencies of continuous and categorical factors and exploiting category-dependent variabilities, the coupled multi-arm architecture enhances accuracy, robustness, and interpretability of the inferred factors without requiring any prior on the relative abundances of categories.

Our contributions can be summarized as follows: (i) We theoretically justify the advantage of the multi-arm VAE framework as a collective decision maker for more accurate inference. (ii) We formulate collective decision making as a variational inference problem with multiple VAE arms and show that this formulation is equivalent to a collection of 
constrained VAEs. The proposed constraint is defined based on the Aitchison geometry in the simplex, which avoids mode collapse. (iii) We benchmark our method and demonstrate its superiority over comparable approaches using multiple datasets including a single cell gene expression dataset, described by an unstructured data matrix of $\sim$20,000 neurons by 5,000 genes. We demonstrate that our method can be used to discover neuronal types as discrete categories and type-specific genes regulating the continuous within-type variability, such as metabolic state or disease state. 

\textbf{Related work.}
There is an extensive body of research on clustering in mixture models~\cite{dilokthanakul2016deep, jiang2017variational, tian2017deepcluster, guo2016boosting, locatello2018boosting}. The idea of improving the clustering performance through seeking a consensus and co-training across multiple observations has been explored in both unsupervised~\cite{monti2003consensus, kumar2011co} and semi-supervised contexts~\cite{blum1998combining}. However, these methods do not consider the underlying continuous variabilities across observations. 
Beyond assigning clusters, in our framework, autoencoding arms seek a consensus at the time of learning mixture representations.

The proposed framework does not need supervision since the individual arms provide a form of prior or weak supervision for each other. In this regard, our paper is related to a body of work that attempts to improve representation learning by using semi-supervised or group-based settings~\cite{bouchacourt2017multi,hosoya2019group,nemeth2020adversarial}. \citet{bouchacourt2017multi} demonstrated a multi-level variational autoencoder (MLVAE) as a semi-supervised VAE by revealing that observations within groups share the same type. \citet{hosoya2019group} and \citet{nemeth2020adversarial} attempted to improve MLVAE by imposing a weaker condition to the grouped data. In recent studies~\cite{shu2019weakly, locatello2020weakly}, a weakly supervised variational setting has been proposed for disentangled representation learning by providing pairs of observations that share at least one underlying factor. They all rely on learning latent variables in continuous spaces, and have been only applied to image datasets with low-dimensional latent space. 

Recent advances in structured variational methods, such as imposing a prior~\cite{ranganath2016hierarchical} or spatio-temporal dependencies~\cite{quiroz2018gaussian} on the latent distribution parameters, allows for scaling to larger dimensions. However, these solutions are not directly applicable to the discrete space, which will be addressed in our A-arm VAE framework.
\section{Single mixture VAE framework}
\label{sec:Single VAE Framework}
For an observation $\mathbf{x} \in \mathbb{R}^D$, a VAE learns a generative model $p_{\boldsymbol{\theta}}\left(\mathbf{x} | \mathbf{z} \right)$ and a variational distribution $q_{\boldsymbol{\phi}}\left(\mathbf{z} | \mathbf{x} \right)$, where $\mathbf{z} \in \mathbb{R}^M$ for $M \ll D$ is a latent variable with a parameterized distribution $p(\mathbf{z})$~\cite{kingma2013auto}. \textit{Disentangling} different sources of variability into different dimensions of $\mathbf{z}$ enables an interpretable selection of latent factors~\cite{higgins2017beta,locatello2018challenging}. However, in many real-world applications, the inherent mixture distribution of continuous and discrete variations is often overlooked. This problem can be addressed within the VAE framework in an unsupervised fashion by introducing a categorical latent variable $\mathbf{c} \in \mathcal{S}^K$, denoting the class label defined in a $K$-simplex, alongside the continuous latent variable $\mathbf{s} \in \mathbb{R}^M$. Here, we refer to the continuous variable $\mathbf{s}$ as the \textit{state} or \textit{style} variable interchangeably. Assuming $\mathbf{s}$ and $\mathbf{c}$ are independent random variables, the evidence lower bound (ELBO)~\cite{blei2017variational} for a single mixture VAE with the distributions parameterized by $\boldsymbol{\theta}$ and $\boldsymbol{\phi}$ is given by,
\begin{multline}
    \label{eq:original_jointVAE}
    \hspace{-.1in} \mathcal{L}(\boldsymbol{\phi}, \boldsymbol{\theta}) = \mathbb{E}_{q_{\boldsymbol{\phi}}(\mathbf{s},\mathbf{c}|\mathbf{x})}\left[\log{p_{\boldsymbol{\theta}}(\mathbf{x}|\mathbf{s},\mathbf{c})}\right] - \\
    D_{KL}\left(q_{\boldsymbol{\phi}}(\mathbf{s}|\mathbf{x}) \| p(\mathbf{s}) \right) 
    -D_{KL}\left(q_{\boldsymbol{\phi}}(\mathbf{c}|\mathbf{x}) \| p(\mathbf{c}) \right)
\end{multline}
\histogram
Maximizing ELBO in Eq.~\ref{eq:original_jointVAE} imposes characteristics on $q(\mathbf{s}|\mathbf{x})$ and $q(\mathbf{c}|\mathbf{x})$ that can result in underestimation of posterior probabilities such as the mode collapse problem, where the network ignores a subset of latent variables~\cite{minka2005divergence,blei2017variational}. Recently, VAE-based solutions were proposed by imposing a uniform structure on $p(\mathbf{c})$: akin to $\beta$-VAE~\cite{higgins2017beta, burgess2018understanding}, JointVAE~\cite{dupont2018learning} modified the ELBO by assigning a pair of controlled information capacities for each variational factor, i.e. $\mathcal{C}_s \in \mathbb{R}^{|\mathbf{s}|}$ and $\mathcal{C}_c \in \mathbb{R}^{|\mathbf{c}|}$. The main drawback of JointVAE is that its performance is tied to heuristic tuning of $|\mathbf{s}| \times |\mathbf{c}|$ capacities over training iterations, which is not easily scalable to high-dimensional settings. CascadeVAE~\cite{jeong2019learning} suggested another VAE-based mixture model that maximizes the ELBO through a semi-gradient-based algorithm by iterating over two separate optimizations for the continuous and categorical variables. Although the computational cost for the additional optimization for the categorical variable has an approximately linear dependence on the number of categories and batch size, it can still be a deterrent for problems with numerous categories and unbalanced datasets requiring larger batch sizes. 
Thus, earlier solutions fall short of efficiently learning interpretable mixture representations for high-dimensional discrete spaces.

In addition to the issues discussed above, the performance and interpretability of those approaches are further limited by the common assumption that the continuous variable representing the style of the data is independent of the categorical variable. In practice, style often depends on the class label. For instance, even for the well-studied MNIST dataset, the histograms of common digit styles, e.g. ``width'', markedly vary for different digits~(Fig.~\ref{fig:histogram}). Moreover, further analysis of the identified continuous factor in the earlier approaches reveals that the independence assumption among $q(\mathbf{s}|\mathbf{x})$ and $q(\mathbf{c}|\mathbf{x})$ can be significantly violated~(see Supplementary G and H).
\vspace{-0.1in}
\section{Coupled mixture VAE framework}
\label{sec:Coupled mixture VAE Model}
\vspace{-0.05in}
\introFigure
The key intuition behind multi-arm networks is cooperation for decision making to improve posterior estimation. Collective decision making has been studied under different contexts and a popular name referring to its advantages is the \textit{``wisdom of the crowd''}~\cite{surowiecki2005wisdom}. When unanimous decisions made by a crowd~(multiple arms) form a probability distribution, multiple estimates from the same sample increase the expected probability of a true assignment. 

{\bf{A-arm VAE Framework}}.
We define the $A$-arm VAE as an $A$-tuple of independent and architecturally identical autoencoding arms, where the $a$-th arm parameterizes a mixture model distribution~(Fig.~\ref{fig:introFigure}a). In this framework, individual arms receive a collection of non-identical copies, i.e. $\{\mathbf{x}_a, \mathbf{x}_b, \ldots \}$ of the given sample, i.e. $\mathbf{x}$, with the same class label.
While each arm has its own mixture representation with potentially non-identical parameters, all arms cooperate to learn $q(\mathbf{c}_a | \mathbf{x}_a)$ via a cost function at the time of training.
Accordingly, a crowd of VAEs with $A$ arms can be formulated as a collection of constrained variational objectives as follows. 
\begin{equation}
\begin{array}{cc}
\label{eq:const_max_prob}
     \max & \mathcal{L}_{\mathbf{s}_1|\mathbf{c}_1}(\boldsymbol{\phi}_1, \boldsymbol{\theta}_1) + \dots + \mathcal{L}_{\mathbf{s}_A|\mathbf{c}_A}(\boldsymbol{\phi}_A, \boldsymbol{\theta}_A) \\[.05in]
    & \text{s.t.} \ \mathbf{c}_1 = \dots = \mathbf{c}_A
\end{array}
\end{equation}
where $\mathcal{L}_{\mathbf{s}_a|\mathbf{c}_a}(\boldsymbol{\phi}_a, \boldsymbol{\theta}_a)$ is the variational loss for arm $a$, 
\begin{multline}
\label{eq:single_mixVAE_ELBO}
    \hspace{-.1in}\mathcal{L}_{\mathbf{s}_a|\mathbf{c}_a}(\boldsymbol{\phi}_a, \boldsymbol{\theta}_a) = \mathbb{E}_{q(\mathbf{s}_a,\mathbf{c}_a|\mathbf{x}_a)}\left[\log{p(\mathbf{x}_a|\mathbf{s}_a,\mathbf{c}_a)} \right] \\
     - \mathbb{E}_{q(\mathbf{c}_a|\mathbf{x}_a)}\left[D_{KL}\left(q(\mathbf{s}_a|\mathbf{c}_a,\mathbf{x}_a) \| p(\mathbf{s}_a|\mathbf{c}_a) \right) \right] \\
    -\mathbb{E}_{q(\mathbf{s}_a|\mathbf{c}_a,\mathbf{x}_a)}\left[D_{KL}\left(q(\mathbf{c}_a|\mathbf{x}_a) \| p(\mathbf{c}_a) \right) \right].
\end{multline}
In Eq.~(\ref{eq:single_mixVAE_ELBO}), the variational loss for each arm is defined according to the graphical model in Fig.~\ref{fig:introFigure}b, which is built upon the traditional ELBO in Eq.~(\ref{eq:original_jointVAE}) by conditioning the continuous state on the categorical variable (derivation in Supplementary Section B).

\textbf{Arms observe non-identical copies of samples.} In the A-arm VAE framework, VAE arms receive non-identical observations that share the discrete variational factor. To achieve this in a fully unsupervised setting, we use \textit{type-preserving} data augmentation that generates independent and identically distributed copies of data while preserving its categorical identity. For image datasets, conventional transformations such as rotation, scaling, or translation can serve as type-preserving augmentations. However, for non-image datasets, e.g. single-cell data, we seek a generative model that learns transformations representing within-class variability in an unsupervised manner.
To this end, inspired by DAGAN~\citep{antoniou2017data} and VAE-GAN~\citep{larsen2016autoencoding}, we develop a generative model to provide collections of observations for our multi-arm framework~(for additional information, see supplementary Section F).

In Supplementary Section A, Remark~2, we further discuss an under-exploration scenario in data augmentation, in which the augmented samples are not independently distributed and are concentrated around the given sample.

The consensus constraint in the $A$-arm framework (Eq.~\ref{eq:const_max_prob}) regularizes the posterior approximation and enhances variational inference accuracy compared to a single VAE. This is theoretically justified by Propositions~\ref{prop:augmentation} and \ref{prop:arm_number}. 
\begin{define}{\bf{(Confidence)}}
\label{define:confidence}
Suppose $\mathbf{x}$ is generated by a multivariate mixture distribution so that $p(\mathbf{x}) =  {\sum_{{c}}} \int_{{s}}\  p(\mathbf{x}|{s},{c}) p({s}|{c}) p({{c}}) \ d{s}$, where $p({c})$ and $p({s}|{c})$ denote arbitrary distributions for discrete and continuous factors, respectively. For samples belonging to category ``$m$'', the assignment confidence for category ``$k$'' can be expressed as,
\begin{equation}
    \label{eq:confidence}
        \mathcal{C}_m(k) = \mathbb{E}_{\mathbf{x}|m}\left[ \log{p( c=k|\mathbf{x})}\right].\\[-.1in]
\end{equation}
\end{define} 
Consequently, $\mathcal{C}_m(m)$ denotes the confidence of the mixture model for the true categorical factor. In the following propositions, we use $\mathcal{C}^{A}_m(\cdot)$ to convey the categorical confidence of the $A$-arm framework, where $A>0$ denotes the number of VAE arms. 
\begin{prop}
    \label{prop:augmentation}
   Consider the problem of mixture representation learning in a multi-arm VAE framework. For $A > B \geq 1$ and $\forall m$,
   \begin{equation}
   \label{eq:prop_augmentation}
\mathcal{C}^A_m(m) > \mathcal{C}^B_m(m).
   \end{equation}~\em{(Proof in Supplementary Section A)}
\end{prop}
\begin{prop}
    \label{prop:arm_number}
   In the $A$-arm VAE framework, there exists an $A$ such that $\forall m,n$, $m \neq n$,
   \begin{equation}
   \label{eq:prop_arm_number}
       \mathcal{C}^A_m(m) > \mathcal{C}^A_m(n),
   \end{equation}
   independent of the relative abundances of categories.~\em{(Proof in Supplementary Section A)}
\end{prop}
Thus, the consensus constraint is sufficient to enhance inference for mixture representations in the $A$-arm VAE framework. 
Our theoretical results show that the minimum number of arms guaranteeing correct assignment is a function of the prior distribution for the categorical variable and category-dependent distribution. While in an unsupervised approach, defining the required number of arms in the absence of the categorical prior and category-dependent information
remains a challenge, we now show that in the particular case of uniformly distributed categories, one pair of coupled arms is enough to satisfy Eq.~\ref{eq:prop_arm_number}.
\begin{coroll}
    \label{coroll:uniform_case}
        For a uniform prior on the discrete factor, one pair of VAE arms ($A=2$) is sufficient to satisfy Eq.~\ref{eq:prop_arm_number}.~\em{(see Supplementary Section A)}
\end{coroll}
We emphasize the joint presence of discrete and continuous factors in these results. Unlike ~\citet{bouchacourt2017multi,shu2019weakly,locatello2020weakly}, the suggested framework is not restricted to the continuous space, and does not require any weak supervision as~\cite{bouchacourt2017multi}. Instead, it relies on representations that are invariant under non-identical copies of observations. 
\subsection{cpl-mixVAE: pairwise coupling in A-arm VAE}
\label{sec:Multi-arm VAE}
In the $A$-arm VAE framework, the mixture representation is obtained through the optimization in Eq.~\ref{eq:const_max_prob}. Not only is it challenging to solve the maximization in Eq.~\ref{eq:const_max_prob} due to the equality constraint, but the objective remains a function of $p(\mathbf{c})$ which is unknown, and typically non-uniform. To overcome this, we use an equivalent formulation for Eq.~\ref{eq:const_max_prob} by applying the pairwise coupling paradigm as follows~(details of derivation in Supplementary Section C).
\begin{eqnarray}
    \max &\!\!\!\!\!\!& \  \displaystyle{\sum_{a=1}^{A}} \
    (A-1)\bigg(\mathbb{E}_{q(\mathbf{s}_a,\mathbf{c}_a|\mathbf{x}_a)}\left[\log{p(\mathbf{x}_a|\mathbf{s}_a,\mathbf{c}_a)} \right] -  \nonumber \\[-.1in]
   &\!\!\!\!\!\!& \hspace{.2in} \mathbb{E}_{q(\mathbf{c}_a|\mathbf{x}_a)}\left[D_{KL}\left(q(\mathbf{s}_a|\mathbf{c}_a,\mathbf{x}_a) \| p(\mathbf{s}_a|\mathbf{c}_a) \right) \right]\bigg) - \nonumber\\[-.05in]
   &\!\!\!\!\!\!& \displaystyle{\sum_{a<b}}\  
    \mathbb{E}_{q(\mathbf{s}_a|\mathbf{c}_a,\mathbf{x}_a)}\mathbb{E}_{q(\mathbf{s}_b|\mathbf{c}_b,\mathbf{x}_b)}\left[\mathcal{D}(a,b) \right] \nonumber\\[-0.05in]
    &\!\!\!\!\!\!& \hspace{.5in} \text{s.t.} \  \mathbf{c}_a = \mathbf{c}_b \ \ \forall a,b \in [1,A], \ a < b
\label{eq:cpl-mixVAE}
\end{eqnarray}
where $\mathcal{D}(a,b) = D_{KL}\left(q(\mathbf{c}_a|\mathbf{x}_a)q(\mathbf{c}_b| \mathbf{x}_b) \| p(\mathbf{c}_a,\mathbf{c}_b) \right)$,
%
%
is the KL divergence across coupled arms, which is a function of the joint distribution $p(\mathbf{c}_a,\mathbf{c}_b)$, rather than $p(\mathbf{c})$.

We relax the optimization in Eq.~\ref{eq:cpl-mixVAE} into an unconstrained problem by marginalizing the joint distribution over a mismatch measure between categorical variables~(full derivation in Supplementary Section D).\\[-0.2in]
\begin{eqnarray}
\label{eq:cpl_mixVAE_approx}
    \max &\!\!\!\!\!\!& \  \displaystyle{\sum_{a=1}^{A}} \
        (A-1)\bigg(\mathbb{E}_{q(\mathbf{s}_a,\mathbf{c}_a|\mathbf{x}_a)}\left[\log{p(\mathbf{x}_a|\mathbf{s}_a,\mathbf{c}_a)} \right] -  \nonumber \\[-.1in]
       &\!\!\!\!\!\!& \hspace{.2in} \mathbb{E}_{q(\mathbf{c}_a|\mathbf{x}_a)}\left[D_{KL}\left(q(\mathbf{s}_a|\mathbf{c}_a,\mathbf{x}_a) \| p(\mathbf{s}_a|\mathbf{c}_a) \right) \right]\bigg) + \nonumber\\[-.05in]
    &\!\!\!\!\!\!& \hspace{-.45in}\displaystyle{\sum_{a<b}}\  
    H(\mathbf{c}_a| \mathbf{x}_a) + H(\mathbf{c}_b | \mathbf{x}_b) - \lambda \mathbb{E}_{q(\mathbf{c}_a, \mathbf{c}_b|\mathbf{x}_a, \mathbf{x}_b)}\left[d^2(\mathbf{c}_a,\mathbf{c}_b) \right] \nonumber \\[-.1in]
\end{eqnarray} \\[-0.3in]
In Eq.~\ref{eq:cpl_mixVAE_approx}, in addition to entropy-based confidence penalties known as mode collapse regularizers~\cite{pereyra2017regularizing}, the distance measure $d(\mathbf{c}_a, \mathbf{c}_b)$ encourages a consensus on the categorical assignment controlled by $\lambda \ge 0$, called \textit{coupling} hyperparameter. 

We refer to the model in Eq.~\ref{eq:cpl_mixVAE_approx} as \textit{cpl-mixVAE}~(Fig.~\ref{fig:introFigure}a). In cpl-mixVAE, VAE arms try to achieve identical categorical assignments while independently learning their own style variables. Here, we set $\lambda=1$ universally, though further optimization is possible. While the bottleneck architecture already encourages interpretable continuous variables, this formulation can be easily extended to include an additional hyperparameter to promote disentanglement of continuous variables as in $\beta$-VAE \cite{higgins2017beta}.

It is also instructive to cast Eq.~\ref{eq:cpl_mixVAE_approx} in an equivalent constrained variational optimization.
\begin{remark}
\label{rem:constrainedVAE}
The A-arm VAE framework is a collection of constrained variational models as follows,
\begin{eqnarray}
\label{eq:coll_single_vae}
    \max &\!\!\!\!\!\!& \  \displaystyle{\sum_{a=1}^{A}} \
        \mathbb{E}_{q(\mathbf{s}_a,\mathbf{c}_a|\mathbf{x}_a)}\left[\log{p(\mathbf{x}_a|\mathbf{s}_a,\mathbf{c}_a)} \right] -  \nonumber \\
       &\!\!\!\!\!\!& \hspace{-0.3in} \mathbb{E}_{q(\mathbf{c}_a|\mathbf{x}_a)}\left[D_{KL}\left(q(\mathbf{s}_a|\mathbf{c}_a,\mathbf{x}_a) \| p(\mathbf{s}_a|\mathbf{c}_a) \right) \right] + H(\mathbf{c}_a|\mathbf{x}_a) \nonumber\\
    &\!\!\!\!\!\!& \hspace{.2in} \text{s.t.} \  \mathbb{E}_{q(\mathbf{c}_a|\mathbf{x}_a)}\left[d^2(\mathbf{c}_a,\mathbf{c}_b) \right] < \epsilon
\end{eqnarray}
where $\epsilon$ denotes the strength of the consensus constraint. Here, $\mathbf{c}_b$ indicates the assigned category by any one of the arms, $b\in \{1,\ldots,A\}$, imposing structure on the discrete variable as an approximation of the prior.
\end{remark}
{\bf{Distance between categorical variables.}} $d(\mathbf{c}_a, \mathbf{c}_b)$ denotes the distance between a pair of $|\mathbf{c}|$-dimensional un-ordered categorical variables, which are associated with probability vectors with non-negative entries and sum-to-one constraint that form a $K$-dimensional simplex, where $K=|\mathbf{c}|$. In the real space, a typical choice to compute the distance between two vectors is using Euclidean geometry. However, this geometry is not suitable for probability vectors. Here, we utilize \textit{Aitchison geometry}~\cite{aitchison1982statistical,egozcue2003isometric}, which defines a substitute vector space on the simplex. 
Accordingly, the distance in the simplex, i.e. $d_{S^K}(\mathbf{c}_{a}, \mathbf{c}_{b})$ is defined as $d_{S^K}(\mathbf{c}_{a}, \mathbf{c}_{b}) = \| clr(\mathbf{c}_{a}) - clr(\mathbf{c}_{b})\|_2$, $\forall \mathbf{c}_a, \mathbf{c}_b \in \mathcal{S}^K$, where $clr(\cdot)$ denotes the \textit{isometric centered-log-ratio} transformation in the simplex. This categorical distance satisfies the conditions of a mathematical metric according to Aitchison geometry.
\subsection{Handshake in the simplex}
\label{sec:Handshake in the simplex}
An instance of the well-known mode collapse problem~\cite{lucas2019don} manifests itself in the minimization of $d_{S^K}(\mathbf{c}_a,\mathbf{c}_b)$ (Eq.~\ref{eq:cpl_mixVAE_approx}): its trivial local optima encourages the network to abuse the discrete latent factor by ignoring many of the available categories. For instance, in an extreme case, the network learns $\mathbf{c}_{a}=\mathbf{c}_{b}=\mathbf{c}_0$. In this scenario, the continuous variable is compelled to act as a primary latent factor, while the model fails to deliver an interpretable mixture representation despite achieving an overall low loss value.

To avoid such undesirable local equilibria while training, we add perturbations to the categorical representation of each arm, where the perturbation of each arm, i.e. $\mathbf{p}_a \in \mathcal{S}^K$ is a function of mini-batch statistics. If posterior probabilities in the simplex have a small dispersion, the perturbed distance calculation overstates the discrepancies. Instead of minimizing $d^2_{S^K}(\mathbf{c}_{a}, \mathbf{c}_{b})$, we minimize a perturbed distance $d^2_{\sigma}(\mathbf{c}_{a},\mathbf{c}_{b}) = \sum_{k}\left(\sigma^{-1}_{a_{k}}\log{c_{a_k}} -  \sigma^{-1}_{b_k}\log{c_{b_k}}\right)^2$, 
where $\sigma^2_{a_k}$ and $\sigma^2_{b_k}$ indicate the mini-batch variance of the $k$-th category, for arms $a$ and $b$. We show that the perturbed distance $d_{\sigma}(\cdot)$ is bounded by $d_{S^K}(\cdot)$ and non-negative values $\rho_u,\rho_l$, which are function of $d_{S^K}(\mathbf{p}_{a}, \mathbf{p}_{b})$.
\begin{prop}
    \label{prop:clr_distance_approx}
     Suppose $\mathbf{c}_{a},\mathbf{c}_{b} \in \mathcal{S}^K$, where $\mathcal{S}^K$ is a simplex of $K > 0$ parts. If  $d_{S^K}\left(\mathbf{c}_{a}, \mathbf{c}_{b} \right)$ denotes the distance in Aitchison geometry and $d^2_{\sigma}(\mathbf{c}_{a},\mathbf{c}_{b}) = \sum_{k}\left(\sigma^{-1}_{a_{k}}\log{c_{a_k}} -  \sigma^{-1}_{b_k}\log{c_{b_k}}\right)^2$ denotes a perturbed distance, then
    \begin{equation*}
    \centering
        d^2_{S^K}\left(\mathbf{c}_{a}, \mathbf{c}_{b} \right) - \rho_l \leq d^2_{\sigma}\left(\mathbf{c}_{a}, \mathbf{c}_{b} \right) \leq  d^2_{S^K}\left(\mathbf{c}_{a}, \mathbf{c}_{b} \right) + \rho_u
    \end{equation*}
    where $\rho_u, \rho_l \geq 0$, $\rho_u=K\left(\tau^2_{{\sigma}_u} + \tau^2_{\mathbf{c}} \right) + 2\Delta_{{\sigma}}\tau_{\mathbf{c}}$, $\rho_l= \displaystyle{\frac{\Delta_{\sigma}^2}{K}} - K\tau^2_{{\sigma}_l}$, $\tau_{\mathbf{c}} = \displaystyle{\max_{k}}\{\log{c_{a_k}} - \log{c_{b_k}}\}$, $\tau_{{\sigma}_u}=\displaystyle{\max_{k}}\{g_k\}$, $\tau_{{\sigma}_l}=\displaystyle{\max_{k}}\{-g_k\}$, $\Delta_{\sigma}= \displaystyle{\sum_k}g_k$, and $g_k=(\sigma^{-1}_{a_k}-1)\log{c_{a_k}} - (\sigma^{-1}_{b_k}-1)\log{c_{b_k}}$.~{\em (Proof in Supplementary Section E)}
\end{prop}
Thus, when $\mathbf{c}_a$ and $\mathbf{c}_b$ are similar and their spread is not small, $d_{\sigma}(\mathbf{c}_a,\mathbf{c}_b)$ closely approximates $d_{S^K}(\mathbf{c}_a,\mathbf{c}_b)$. Otherwise, it diverges from $d_{S^K}(\cdot)$ to avoid mode collapse.
\vspace{-0.1in}
\section{Experiments}
\label{sec:Experiments}
\vspace{-0.05in}
%
\allResults
We used three datasets: MNIST, dSprites, and a single-cell RNA-sequencing dataset (scRNA-seq)~\cite{tasic2018shared}. For the scRNA-seq dataset, identifying interpretable biological variables is a challenge: neuronal cells, the basic building blocks of the brain, display both significant diversity and stereotypy in their gene expression. Individual cells can differ due to either their \textit{type} or continuous \textit{within-type} variations~\cite{trapnell2015typestate, andrews2018identifying}, which are considered as biological interpretations of discrete and continuous variabilities. Identifying such factors can be useful to study canonical brain circuits in terms of their \textit{generic} components~\cite{bargmann2014brain}, and to identify gene expression programs~\cite{trapnell2015typestate}, both of which are high-priority research areas in neuroscience.

Although MNIST and dSprites datasets do not require high-dimensional settings for mixture representation, to facilitate comparisons of cpl-mixVAE with earlier methods, first we report the results for these benchmark datasets. 

We trained three unsupervised VAE-based methods for mixture modeling:~JointVAE~\cite{dupont2018learning}, CascadeVAE~\cite{jeong2019learning}, and ours~(cpl-mixVAE). For  MNIST, we additionally trained the popular InfoGAN~\cite{chen2016infogan} as the most comparable GAN-based model. To report the interpretability of the mixture representations, we consider the accuracy~(ACC) of categorical assignments for the discrete variable and latent traversal analysis for the continuous variable by fixing the discrete factor and changing the continuous variable according to $p(\mathbf{s}|\mathbf{c},\mathbf{x})$. Additionally, we report the computational efficiency~(number of iterations per second) to compare the training complexity of the multi-arm framework against earlier methods~(Table~\ref{tab:allResults}). All reported numbers for cpl-mixVAE models are average accuracies calculated across arms.

During training of VAE-based models, to sample from $q(\mathbf{c}_a|\mathbf{x}_a)$, we use the Gumbel-softmax distribution~\citep{jang2016categorical,maddison2014sampling} with a temperature parameter $0 < \tau \leq 1$. In cpl-mixVAE, each arm received an augmented copy of the original input generated by the deep generative augmenter~(Supplementary Section F), while training. Details of the networks architectures and training settings can be found in Supplementary Section J. 
\vspace{-0.1in}
\subsection{Benchmark datasets}
\vspace{-0.05in}
{\bf{MNIST.}} Based on the uniform distribution of handwritten digits in MNIST, we used a 2-arm cpl-mixVAE to learn interpretable representations. Following the convention~\cite{chen2016infogan,dupont2018learning, jeong2019learning}, each arm of cpl-mixVAE uses a 10-dimensional categorical variable representing digits (type), and a 10-dimensional continuous random variable representing the writing style (state).
Table~\ref{tab:allResults} displays the classification performance of the discrete latent variable (the predicted class label) for InfoGAN, two 1-arm VAE methods (JointVAE and CascadeVAE), and cpl-mixVAE with 2 arms. Additionally, Fig.~\ref{fig:allDataLatentTraversa}~(top panel) illustrates the continuous latent traversals for four dimensions of the state variable inferred by cpl-mixVAE, where each row corresponds to a different dimension of the categorical variable, and the state variable monotonically changes across columns. 
Both results in Table~\ref{tab:allResults} and Fig.~\ref{fig:allDataLatentTraversa} show that cpl-mixVAE achieved an interpretable mixture representation with the highest categorical assignment accuracy.

\textbf{dSprites.} Similarly, due to the uniform distribution of classes, we again used a 2-arm cpl-mixVAE model. Results in Table~\ref{tab:allResults} verifies that our method outperforms the other methods in terms of categorical assignment accuracy. To report the intractability of the continuous variable, in addition to demonstrating the traversal results~(Fig.~\ref{fig:allDataLatentTraversa}~(bottom panel)), we reported disentanglement scores. For a fair comparison, we used the same disentanglement metric implemented for CascadeVAE~\cite{jeong2019learning}.

\textbf{Summary.} cpl-mixVAE improves the accuracy of categorical assignments and infers better mixture representations. It outperforms earlier methods, without using extraneous optimization or heuristic channel capacities. Beyond performance and robustness, its computational cost is also comparable to that of the baselines. 
\allDataLatentTraversa
\vspace{-0.1in}
\subsection{scRNA-seq} 
\vspace{-0.1in}
scRNA-seq datasets are significantly more complex than a typical machine learning dataset. Here, the scRNA-seq dataset includes transcriptomic profiles of $10,000$ genes for $22,365$ cells, from over $100$ cell types with sizeable differences between the relative abundances of clusters. Hence, two main challenges of representation learning in this dataset are (i) large number of cell types~(discrete variable), and (ii) class imbalance -- the most- and the least-abundant cell types include $1404$ and $16$ samples, respectively. Moreover, whether the observed diversity corresponds to discrete variability or a continuum is an ongoing debate in neuroscience.  
While using genes that are differentially expressed in subsets of cells, known as \textit{marker genes}~(MGs)~\cite{trapnell2015typestate} is a common approach to define cell types, the identified genes rarely obey the idealized MG definition in practice. Identifying these biomarkers is a challenging process, since highest variance genes are often expressed and may be known markers for multiple cell types. 
Here, we focus on \textit{neuronal} cells and use a subset of $5,000$ highest variance genes. The original neuroscience study suggested $115$ discrete neuronal types, based on an MG-based approach~\cite{tasic2018shared}.

\textbf{Neuron type identification.} We trained mixture VAE-based models using 115- and 2-dimensional discrete and continuous variables. We compared the suggested cell types in~\cite{tasic2018shared} with the discrete representations that are inferred from VAE models. Table~\ref{tab:allResults} and Fig.~\ref{fig:clusteringFACSall} demonstrate the performance of a 2-arm cpl-mixVAE model against JointVAE and CascadeVAE. Our results clearly show that cpl-mixVAE outperforms JointVAE and CascadeVAE in identifying meaningful known cell types.

\textbf{Using $\mathbf{A>2}$.} Unlike the discussed benchmark datasets, the neuronal types are not uniformly distributed. Accordingly, we also investigated the accuracy improvement for categorical assignment when more than two arms are used. Fig.~\ref{fig:multiagentFACS} illustrates the accuracy improvement with respect to a single autoencoder model, i.e. JointVAE, in agreement with our theoretical findings.

\clusteringFACSall
\textbf{Identifying genes regulating cell activity.} To examine the role of the continuous latent variable, 
we applied a similar traversal analysis to that used for the benchmark datasets. For a given cell sample and its discrete type, we changed each dimension of the continuous variable using the conditional distribution, and inspected gene expression changes caused by continuous variable alterations. Fig.~\ref{fig:latentTraversalFACS} shows the results of the continuous traversal study for a 1-arm VAE~(JointVAE) and a 2-arm VAE~(cpl-mixVAE), for two excitatory neurons belonging to the ``L5 NP''~(cell type~(I)) and ``L6 CT''~(cell type~(II)) sub-classes. In each sub-figure, the latent traversal is color-mapped to normalized reconstructed expression values, where the $y$-axis corresponds to one dimension of the continuous variable, and the $x$-axis corresponds to three gene subsets, namely (i) MGs for the two excitatory types, (ii) immediate early genes (IEGs), and (iii) housekeeping gene (HKGs) subgroups~\cite{hrvatin2018single, tarasenko2017cytochrome}. For cpl-mixAVE~(Fig.~\ref{fig:latentTraversalFACS}b), the normalized expression of the reported MGs as indicators for excitatory cell types (discrete factors) is unaffected by changes of identified continuous variables. In contrast, for JointVAE~(Fig.~\ref{fig:latentTraversalFACS}a), we observed that the normalized expression of some MGs~(5 out of 10) are changed due to the continuous factor traversal. Additionally, we found that the expression changes inferred by cpl-mixVAE for IEGs and HKGs are essentially \textit{monotonically} linked to the continuous variable, confirming that the expression of IEGs and HKGs depends strongly on the cell activity variations under different metabolic and environmental conditions. Conversely, JointVAE fails to reveal such activity-regulated monotonicity for IEGs and HKGs. Furthermore, our results for cpl-mixVAE reveal that the expression of activity-regulated genes depends on the cell type, i.e. IEGs and HKGs respond differently to activation depending on their cell types~(compare rows I and II in Fig.~\ref{fig:latentTraversalFACS}b). However, in Fig.~\ref{fig:latentTraversalFACS}a, since the baseline 1-arm VAE does not take into account the dependency of discrete and continuous factors, it fails to reveal the dependence of activity-regulated expression to the cell type, and therefore produces identical expressions for both types (I) and (II).
%
%
\multiagentFACS

\textbf{Summary.} While both JointVAE and CascadeVAE failed to identify cell types as discrete factors, cpl-mixVAE successfully identified the majority of known types. Our findings suggest that cpl-mixVAE  by acknowledging the dependencies of continuous and categorical factors, captures relevant and interpretable continuous variability that can provide insight when deciphering the molecular mechanisms shaping the landscape of biological states, e.g. metabolic or disease.
\vspace{-0.2in}
\subsection{Ablation studies}
To elucidate the success of the A-arm VAE framework in mixture modeling, we investigate the categorical assignment performance under different training settings. Since CascadeVAE does not learn the categorical factors by variational inference, here we mainly study JointVAE (as a 1-arm VAE) and cpl-mixVAE (as a 2-arm VAE). First, to isolate the impact of data augmentation in training, we trained JointVAE$^\dagger$, where the JointVAE model was trained with noisy copies of the original MNIST dataset, same as cpl-mixVAE. The results in Table~S1 (see Supplementary Section I) for JointVAE$^\dagger$ suggest that data augmentation by itself does not enhance the categorical assignment. Subsequently, to understand whether architectural differences put JointVAE at a disadvantage, we trained JointVAE$^\ddagger$~(Table~S1), which uses the same architecture as the one used in cpl-mixVAE.  JointVAE$^\ddagger$ uses the same learning procedure as JointVAE, but its convolutional layers are replaced by fully-connected layers (see Supplementary Section I and J for details). The result for JointVAE$^\ddagger$ suggests that the superiority of cpl-mixVAE is not due to the network architecture either. 
Furthermore, we examined the performance changes of the proposed 2-arm cpl-mixVAE under three different settings: (i) cpl-mixVAE$^*$, where coupled networks are not independent and network parameters are shared; (ii) cpl-mixVAE$^a$, where only random rotations ($[-\pi/9,\pi/9]$) are used as an affine transformation for data augmentation; and (iii) cpl-mixVAE$(\mathbf{s} \centernot\mid \mathbf{c})$, where the state variable is independent of the discrete variable~(Table~S1). Our results show that the proposed cpl-mixVAE obtained the best categorical assignments across all training settings. 

\textbf{Summary.} We experimentally observe that inference does not improve in 1-arm VAEs by using either augmented copies or cpl-mixVAE's single network design. Additionally, when within-class variations can be guessed (image datasets), using a simple augmentation strategy, e.g. cpl-mixVAE$^a$, is sufficient for the A-arm VAE framework. 
\latentTraversalFACS

\vspace{-0.1in}
\section{Conclusion}
\label{sec:Conclusion}
\vspace{-0.1in}
We have proposed cpl-mixVAE as a multi-arm framework to apply the power of collective decision making in unsupervised joint representation learning of discrete and continuous factors, scalable to the high-dimensional discrete space. This framework utilizes multiple pairwise-coupled autoencoding arms with a shared categorical variable, while independently learning the continuous variables. Our experimental results for all three datasets support the theoretical findings, and show that cpl-mixVAE outperforms comparable models. Importantly, for a challenging scRNA-seq dataset, we showed that the proposed framework can identify biologically interpretable cell types and differentiate between type-dependent and activity-regulated genes.

\bibliography{reference}
\bibliographystyle{icml2021}
\end{document}


{\centering\section*{\huge{Supplementary Materials}}}
\vspace{0.1in}
\appendix
\section{Collective decision making through multi-arm settings}
\begin{prop}
    \label{prop:augmentation}
   Consider the problem of mixture representation learning in a multi-arm VAE framework. For $A > B \geq 1$ and $\forall m$,
   \begin{equation}
   \label{eq:prop_augmentation}
\mathcal{C}^A_m(m) > \mathcal{C}^B_m(m).
   \end{equation}
\end{prop}
\begin{proof}
Following Definition~1 of the multi-arm framework, each arm is represented by 
\begin{equation}
    p(\mathbf{x}_a|c_a,s_a) \propto \frac{p(s_a |c_a, \mathbf{x}_a) p(c_a |\mathbf{x}_a)}{p(s_a|c_a) p(c_a)} \ .
\end{equation}
%
In this framework, using a type-preserving data augmentation, each arm receives a noisy copy $\mathbf{x}_a$ of given sample $\mathbf{x}$. Without loss of generality, let $\mathbf{x} \sim p(\mathbf{x}|m)$, where $m \in \{1,\ldots,K\}$ denotes the true categorical assignment, respectively. Accordingly, $\mathbf{x}_a \sim p(\mathbf{x}_a|m)$ also belong to the same category-conditioned distribution $p(\mathbf{x}|m)$, $\forall m \in \{ 1, \dots, K\}$. Considering the joint categorical assignment as $c$, where $c=c_1=\dots =c_A$. Defining $\mathbf{X}:=\{\mathbf{x}_a\}_{1:A}$, for an $A$-arm VAE, the confidence measure for category $k$ for samples from category $m$ is expressed as,
\begin{eqnarray}
\label{eq:conf_Aarm_0}
     \mathcal{C}^A_{m}(k) &=& \mathbb{E}_{p(\mathbf{X}|m)}\left[\log{p(c=c_1=\dots =c_A=k|\mathbf{X})}\right], \\[.1in] \nonumber
    &=& \mathbb{E}_{p(\mathbf{X}|m)}\left[\log{\displaystyle{\frac{p(\mathbf{x}_1|\mathbf{X} \setminus \{\mathbf{x}_1 \}, c=k)p(\mathbf{x}_2|\mathbf{X} \setminus \{\mathbf{x}_1, \mathbf{x}_2 \}, c=k)\dots p(\mathbf{x}_A|c=k)p(c=k)}{p(\mathbf{X})}}}\right],\\[0.1in] \nonumber
    &=& \mathbb{E}_{p(\mathbf{X}|m)}\left[\log{\displaystyle{\frac{p(\mathbf{x}_1|\mathbf{X} \setminus \{\mathbf{x}_1 \}, c=k)\dots p(\mathbf{x}_A|c=k)}{p(\mathbf{X})}}}\right] + \log{p(c=k)}
\end{eqnarray}
where we used $p(c)=p(c=c_1=\dots =c_A)$ to simplify the notation. Utilizing the type-preserving data augmentation, since all samples independently generated from the same class-conditioned distribution, i.e. $p(\mathbf{x}_a|c, \mathbf{x}_b)=p(\mathbf{x}_a|c)$, the confidence measure can be simplified as,
\begin{eqnarray}
\label{eq:conf_Aarm}
     \mathcal{C}^A_{m}(k) &=& \mathbb{E}_{p(\mathbf{x}_a|m)}\left[\displaystyle{\log{ \prod_{a=1}^A\frac{p(\mathbf{x}_a|c=k)}{p(\mathbf{x}_a)}}}\right] + \log{p(c=k)} \\
     &=&\sum_{a=1}^A \mathbb{E}_{p(\mathbf{x}_a|m)}\left[\displaystyle{ \log{ \frac{p(\mathbf{x}_a|c=k)}{p(\mathbf{x}_a)}}}\right] + \log{p(c=k)}
\end{eqnarray}
Since all of the augmented data are sampled from the same distribution, log-likelihood values of the augmented samples are equal on expectation, as follows:
\begin{eqnarray}
\label{eq:collective_conf}
    \displaystyle{\sum_{a=1}^A}\ \mathbb{E}_{p(\mathbf{x}_a|m)}\left[\log{p(\mathbf{x}_a|c)}\right] = A \mathbb{E}_{p(\mathbf{x}|m)}\left[\log{p(\mathbf{x}|c)}\right]
\end{eqnarray}
Therefore, the confidence over category $k$ in an $A$-arm framework is defined as,
\begin{eqnarray}
\label{eq:conf_Aarm_2}
     \mathcal{C}^A_{m}(k) = A \mathbb{E}_{p(\mathbf{x}|m)}\left[\log{\displaystyle{\frac{p(\mathbf{x}|c=k)}{p(\mathbf{x})}}}\right] + \log{p(c=k)}.
\end{eqnarray}
According to Eq.~\ref{eq:conf_Aarm_2}, for a single arm (1-arm) and $A$-arm frameworks, the confidence values for the true categorical assignment, i.e. category $m$, are formulated as follows.
\begin{eqnarray}
\label{eq:conf_1arm_m}
     \mathcal{C}^1_m(m) &=&  \mathbb{E}_{p(\mathbf{x}|m)}\left[\log{\displaystyle{\frac{p(\mathbf{x}|c=m)}{p(\mathbf{x})}}}\right] + \log{p(c=m)}, \\ \nonumber
     &=& D_{KL}\left(p(\mathbf{x}|c=m) \| p(\mathbf{x})  \right) + \log{p(c=m)}\\ [0.15in]
     \mathcal{C}^A_m(m) &=& A\, D_{KL}\left(p(\mathbf{x}|c=m) \| p(\mathbf{x})  \right) + \log{p(c=m)}, \label{eq:conf_1arm_m_A}
\end{eqnarray}
Since $D_{KL}\left(p(\mathbf{x}|c=m) \| p(\mathbf{x})\right) > 0$ for $K > 1$, for $A > B \geq 1$ and $\forall m$, we have $\mathcal{C}^A_m(m) > \mathcal{C}^B_m(m)$.\\
%
%
\end{proof}
\begin{prop}
    \label{prop:arm_number}
   In the $A$-arm VAE framework, there exists an $A$ such that $\forall m,n$, $m \neq n$,
   \begin{equation}
   \label{eq:prop_arm_number}
       \mathcal{C}^A_m(m) > \mathcal{C}^A_m(n),
   \end{equation}
   independent of the relative abundances of categories.
\end{prop}
\begin{proof}
In the A-arm framework, the joint assignment receives the highest confidence score, if and only if,
\begin{eqnarray}
\label{eq:joint_confidence}
     \mathcal{C}^A_m(m) &>& \mathcal{C}^A_m(n),  \ \ \forall  n \neq m \nonumber 
\end{eqnarray}
Considering $\forall \ m,n \in \{1,\ldots,K\}, n \neq m$, $D_{kl}\left(p(c=m|\mathbf{x}) \| p(c=n|\mathbf{x})\right) > 0$, the correct categorical assignment receives the highest confidence for the 1-arm case, i.e. $\mathcal{C}^1_m(m) > \mathcal{C}^1_m(n)$, if and only if,
\begin{eqnarray}
\label{eq:need_to_satisify}
     \mathbb{E}_{p(\mathbf{x}|m)}\left[\log{\displaystyle{\frac{p(\mathbf{x}|c=m)}{p(\mathbf{x}|c=n)}}}\right] &>& \log{\displaystyle{\frac{p(c=n)}{p(c=m)}}},  \ \ \forall  n \neq m 
\end{eqnarray}
which is a function of categorical distributions and is not always satisfied for any arbitrary discrete prior distribution. 
When there are $A$ arms receiving type-preserving noisy copies of the given sample $\mathbf{x}$,
\begin{eqnarray}
\label{eq:collective_conf_2}
     \mathcal{C}^A_m(k)&=& A \mathbb{E}_{p(\mathbf{x}|m)}\left[\log{p(\mathbf{x}|c=k)}\right]+ \log{p(c=k)} - A\mathbb{E}_{p(\mathbf{x}|m)}\left[\log{p(\mathbf{x})}\right],
\end{eqnarray}
Therefore, $\mathcal{C}^A_m(m) > \mathcal{C}^A_m(n),  \ \ \forall  n \neq m$, if and only if,
\begin{equation}
\label{eq:loglikelihood_ratio}
      A\mathbb{E}_{p(\mathbf{x}|m)}\left[\log{\displaystyle{\frac{p(\mathbf{x}|c=m)}{p(\mathbf{x}|c=n)}}}\right] > \log{\displaystyle{\frac{p(c=n)}{p(c=m)}}}, \ \ \forall  n \neq m. 
\end{equation}
Thus, when the number of arms, $A$, satisfies
\begin{eqnarray}
\label{eq:min_num_arms}
      A &>& \displaystyle{\max_{m}}\{\max{(\rho(m) D^{-1}(m), 1)}\},
\end{eqnarray}
where $\rho(m) = \displaystyle{\max_{n \neq m}}\log{\displaystyle{\frac{p(c=n)}{p(c=m)}}}$ and $D(m)=\displaystyle{\min_{n \neq m}} \ D_{KL}(p(\mathbf{x}|m) \| p(\mathbf{x}|n)))$, we have
\begin{equation}
    \mathcal{C}^A_m(m) > \mathcal{C}^A_m(n),  \ \ \forall  n \neq m \ .
\end{equation}
\end{proof}
\begin{coroll}
    \label{coroll:uniform_case}
        For a uniform prior on the discrete factor, one pair of VAE arms ($A=2$) is sufficient to satisfy Eq.~\ref{eq:prop_arm_number}.
\end{coroll}
\begin{proof}
For uniformly distributed clusters, $\rho(k)=0$, $\forall k \in \{1,\dots,K\}$. According to Eq.~\ref{eq:min_num_arms}, for any $A \geq 2$, the confidence increase criteria is satisfied. 
\end{proof}
\noindent
We further study the under-exploration scenario in data augmentation, in which the augmenter does not generate samples with $p(\mathbf{x}|\mathbf{c})$; rather, they are concentrated around the given sample. Under this scenario, the proof of earlier Propositions follow in the same way except the augmented samples are no longer conditionally independent. This means that each augmented sample adds less than before to the confidence score. Yet, the same argument shows that there will be an $A$ for which the claims of the proposition are satisfied. This issue is discussed in Remark~\ref{remark:underexploration} as follows.
\stepcounter{remark}
\begin{remark}
    \label{remark:underexploration}
    When the augmentation is type-preserving, by definition, $p(\mathbf{x}_a|\mathbf{x}_b,c=k)=p(\mathbf{x}_a|c=k)$, where $\mathbf{x}_b$ could be either the given training sample or another noisy copy. If the augmented samples concentrate around $\mathbf{x}_b$, i.e. the augmenter under-explores the category-conditioned distribution, the earlier proofs should be adapted by keeping the conditioning on $\mathbf{x}_b$ explicit. Conditionally independent terms used in Eq.~\ref{eq:conf_Aarm} should be replaced by $p(\mathbf{x}_a|\mathbf{x}_b,c=k)$ as follows.
    \begin{eqnarray}
    \label{eq:conf_1arm_conditioned}
    \mathcal{C}^A_m(k) &=& \mathbb{E}_{p(\mathbf{X}|m)}\left[\log{p(c=c_1=\dots =c_A=k|\mathbf{X})}\right],  \\[0.1in] \nonumber 
    &=& \mathbb{E}_{p(\mathbf{X}|m)}\left[\log{\displaystyle{\frac{p(\mathbf{x}_1|\mathbf{x}_b,\mathbf{X} \setminus \{\mathbf{x}_1, \mathbf{x}_b\}, c=k)\dots p(\mathbf{x}_{A}|\mathbf{x}_b,c=k)p(\mathbf{x}_b|c=k)p(c=k)}{p(\mathbf{X})}}}\right]
    \end{eqnarray}
    Since all augmented samples are generated from sample $\mathbf{x}_b$, the conditional probability distribution can be simplified as follows.
    \begin{equation}
        p(\mathbf{x}_a|\mathbf{x}_b,\mathbf{X} \setminus \{\mathbf{x}_a, \mathbf{x}_b\}, c)=p(\mathbf{x}_a|\mathbf{x}_b,c) , \ \ for \ a \neq b
    \end{equation}
    Accordingly, Eq.~\ref{eq:conf_1arm_conditioned} can be simplified as,
    \begin{eqnarray}
    \label{eq:conf_1arm_simp}
    \mathcal{C}^A_m(k) 
    &=& \mathbb{E}_{p(\mathbf{x}_a|\mathbf{x}_b,m)}\left[\log{\displaystyle \prod_{a=1}^{A-1}{\frac{p(\mathbf{x}_a|\mathbf{x}_b, c=k)}{p(\mathbf{x}_a|\mathbf{x}_b)}}}\right] + \mathbb{E}_{p(\mathbf{x}_b|m)}\left[\log{\displaystyle {\frac{p(\mathbf{x}_b| c=k)}{p(\mathbf{x}_b)}}}\right] + \log{p(c=k)} \\[0.1in] \nonumber
    &=& (A-1)\mathbb{E}_{p(\mathbf{x}_a|\mathbf{x}_b,m)}\left[\log{\displaystyle {\frac{p(\mathbf{x}_a|\mathbf{x}_b, c=k)}{p(\mathbf{x}_a|\mathbf{x}_b)}}}\right] + \mathcal{C}^1_m(k)
    \end{eqnarray}
    Based on Eq.~\ref{eq:conf_1arm_simp}, if the data augmenter only regenerates given sample $\mathbf{x}$, the confidence value is equal to the confidence of the single framework.
    Then, Eq.~\ref{eq:conf_1arm_m_A} should read as
    \begin{eqnarray}
    \label{eq:conf_1arm_conditioned_m}
    \mathcal{C}^A_m(m) &=& (A-1) D_{KL}\left(p(\mathbf{x}_a|\mathbf{x}_b,c=m) \| p(\mathbf{x}_a|\mathbf{x}_b)  \right) + \mathcal{C}^1_m(m) \ .
    \end{eqnarray}
\end{remark}
\noindent
According to Remark~\ref{remark:underexploration}, $\forall \mathbf{x}_a$, if $\mathbf{x}_a = \mathbf{x}_b$, then $ \mathcal{C}^A_m(m) =  \mathcal{C}^1_m(m)$.
\section{Variational lower bound for conditional single mix-VAE}
\label{sec:Single mix-VAE}
For completeness, we first derive the evidence lower bound (ELBO) for an observation $\mathbf{x}$ described by one categorical random variable (RV), $\mathbf{c}$, and one continuous RV, $\mathbf{s}$, without assuming conditional independence of $\mathbf{c}$ and $\mathbf{s}$ given $\mathbf{x}$. The variational approach to choosing the latent variables corresponds to solving the optimization equation
\begin{equation}
    \label{eq:VAE_optim}
    \begin{array}{ll}
    q^{*}(\mathbf{s},\mathbf{c}|\mathbf{x}) = \displaystyle{{\arg\min}_{q(\mathbf{s},\mathbf{c}|\mathbf{x}) \in \mathcal{D}}} \  D_{\mathrm{KL}}\left(q(\mathbf{s},\mathbf{c}|\mathbf{x}) \| p(\mathbf{s},\mathbf{c}|\mathbf{x}) \right)
    \end{array},
\end{equation}
where $\mathcal{D}$ is a family of density functions over the latent variables. However, evaluating the objective function requires knowledge of $p(\mathbf{x})$, which is usually unknown. Therefore, we rewrite the divergence term as 
\begin{eqnarray}
    D_{\mathrm{KL}}\left(q(\mathbf{s},\mathbf{c}|\mathbf{x}) || p(\mathbf{s},\mathbf{c}|\mathbf{x}) \right) &=& \displaystyle{\int_\mathbf{s}} \displaystyle{\sum_{\mathbf{c}}} \ q(\mathbf{s}|\mathbf{c},\mathbf{x})q(\mathbf{c}|\mathbf{x}) \log{\displaystyle{\frac{q(\mathbf{s}|\mathbf{c},\mathbf{x})q(\mathbf{c}|\mathbf{x})}{\displaystyle{\frac{p(\mathbf{x}|\mathbf{s},\mathbf{c})p(\mathbf{s}|\mathbf{c})p(\mathbf{c})}{p(\mathbf{x})}}}}} \ d\mathbf{s} \nonumber\\
    &=& \displaystyle{\int_\mathbf{s}} \displaystyle{\sum_{\mathbf{c}}} \ q(\mathbf{s}|\mathbf{c},\mathbf{x})q(\mathbf{c}|\mathbf{x}) \log{\displaystyle{\frac{q(\mathbf{s}|\mathbf{c},\mathbf{x})}{p(\mathbf{s}|\mathbf{c})}}} \ d\mathbf{s} +\displaystyle{\int_\mathbf{s}} \displaystyle{\sum_{\mathbf{c}}} \ q(\mathbf{s}|\mathbf{c},\mathbf{x})q(\mathbf{c}|\mathbf{x}) \log{\displaystyle{\frac{q(\mathbf{c}|\mathbf{x})}{p(\mathbf{c})}}} \ d\mathbf{s} \nonumber\\
    && +\displaystyle{\int_\mathbf{s}} \displaystyle{\sum_{\mathbf{c}}} \ q(\mathbf{s}|\mathbf{c},\mathbf{x})q(\mathbf{c}|\mathbf{x}) \log{p(\mathbf{x})} \ d\mathbf{s} -\displaystyle{\int_\mathbf{s}} \displaystyle{\sum_{\mathbf{c}}} \ q(\mathbf{s}|\mathbf{c},\mathbf{x})q(\mathbf{c}|x) \log{p(\mathbf{x}|\mathbf{s},\mathbf{c})} \ d\mathbf{s} \nonumber\\
    &=& \log{p(\mathbf{x})} - \mathbb{E}_{q(\mathbf{c}|\mathbf{x})}\left[\mathbb{E}_{(q(\mathbf{s}|\mathbf{c},\mathbf{x}))}\left[\log{p(\mathbf{x}|\mathbf{s},\mathbf{c})} \right] \right] \nonumber\\
    && +\mathbb{E}_{q(\mathbf{c}|\mathbf{x})}\left[D_{KL}\left(q(\mathbf{s}|\mathbf{c},\mathbf{x}) \| p(\mathbf{s}|\mathbf{c}) \right) \right] + \mathbb{E}_{q(\mathbf{s}|\mathbf{c},\mathbf{x})}\left[D_{KL}\left(q(\mathbf{c}|\mathbf{x}) \| p(\mathbf{c}) \right) \right] \label{eq:mixVAE_KLdist}\\
    &=& \log{p(\mathbf{x})} - \mathcal{L}_\mathbf{s} \label{eq:log_likelihood}
\end{eqnarray}
Since $\log{p(\mathbf{x})}$ is unknown, instead of minimizing Eq.~\ref{eq:mixVAE_KLdist}, the variational lower bound
\begin{equation}
\label{eq:single_cond_vae}
    \mathcal{L}_\mathbf{s} = \mathbb{E}_{q(\mathbf{c}|\mathbf{x})}\left[\mathbb{E}_{(q(\mathbf{s}|\mathbf{c},\mathbf{x}))}\left[\log{p(\mathbf{x}|\mathbf{s},\mathbf{c})} \right] \right] - \mathbb{E}_{q(\mathbf{c}|\mathbf{x})}\left[D_{KL}\left(q(\mathbf{s}|\mathbf{c},\mathbf{x}) \| p(\mathbf{s}|\mathbf{c}) \right) \right] 
     -\mathbb{E}_{q(\mathbf{s}|\mathbf{c},\mathbf{x})}\left[D_{KL}\left(q(\mathbf{c}|\mathbf{x}) \| p(\mathbf{c}) \right) \right] 
\end{equation}
can be maximized. We choose $q(\mathbf{s}|\mathbf{c},\mathbf{x})$ to be a factorized Gaussian, parametrized using the reparametrization trick, and assume that the corresponding prior distribution is also a factorized Gaussian, $\mathbf{s}|\mathbf{c} \sim \mathcal{N}(0, \mathbf{I})$. Similarly, for the categorical variable, we assume a uniform prior, $\mathbf{c} \sim U(K)$.
\section{Variational inference for multi-arm autoencoding networks}
\label{sec:Variational inference for multi-arm autoencoding networks}
As discussed in the main text, the collective decision making for an A-arm VAE network can be formulated as an equality constrained optimization as follows.
\begin{equation}
\begin{array}{cc}
\label{eq:max_emc}
     \max & \mathcal{L}(\boldsymbol{\phi}_1, \boldsymbol{\theta}_1,\mathbf{x}_1,\mathbf{s}_1,\mathbf{c}_1) + \dots + \mathcal{L}(\boldsymbol{\phi}_A, \boldsymbol{\theta}_A,\mathbf{x}_A,\mathbf{s}_A,\mathbf{c}_A) \\[.1in]
    & \text{s.t.} \ \mathbf{c}_1 = \dots = \mathbf{c}_A
\end{array}
\end{equation}
Without loss of generality, the optimization in Eq.~\ref{eq:max_emc} can be rephrased as follows.
\begin{equation}
\begin{array}{cc}
     \max & \ \mathcal{L}(\boldsymbol{\phi}_1, \boldsymbol{\theta}_1, \mathbf{s}_1, \mathbf{c}_1) + \mathcal{L}(\boldsymbol{\phi}_2, \boldsymbol{\theta}_2, \mathbf{s}_2, \mathbf{c}_2) + \dots + \mathcal{L}(\boldsymbol{\phi}_A,\boldsymbol{\theta}_A, \mathbf{s}_A, \mathbf{c}_A) \\[.1in]
     & \text{s.t.}  \ \mathbf{c}_1 = \mathbf{c}_2 \\
     & \hspace{.2in}\ \mathbf{c}_1 = \mathbf{c}_3 \\
     & \hspace{.2in}\ \dots \\
     & \hspace{.2in}\ \mathbf{c}_1 = \mathbf{c}_A \\
     & \hspace{.2in}\ \dots \\
     & \hspace{.2in}\ \mathbf{c}_{A-1} = \mathbf{c}_A \\
\end{array}
\end{equation}
where the equality constraint is represented as $\binom{A}{2}$ pairs of categorical agreements. Multiplying the objective term in Eq.~\ref{eq:max_emc} by a constant value, $A-1$, we obtain,
\begin{equation}
\label{eq:modify_max_emc}
\begin{array}{cc}
     \max & \ \left(A-1\right)\left(\mathcal{L}\left(\boldsymbol{\phi}_1, \boldsymbol{\theta}_1, \mathbf{s}_1, \mathbf{c}_1\right) + \mathcal{L}\left(\boldsymbol{\phi}_2, \boldsymbol{\theta}_2, \mathbf{s}_2, \mathbf{c}_2\right) + \dots + \mathcal{L}\left(\boldsymbol{\phi}_A,\boldsymbol{\theta}_A, \mathbf{s}_A, \mathbf{c}_A\right)\right) \\[.1in]
     & \text{s.t.} \ \ \mathbf{c}_a = \mathbf{c}_b \ \ \ \forall a,b \in [1,A], a < b 
\end{array}
\end{equation}
Consider one pair of $\mathcal{L}$ objectives for two arms $a$ and $b$:
\begin{multline}
    \mathcal{L}(\boldsymbol{\phi}_a, \boldsymbol{\theta}_a, \mathbf{s}_a, \mathbf{c}_a) + \mathcal{L}(\boldsymbol{\phi}_b, \boldsymbol{\theta}_b, \mathbf{s}_b, \mathbf{c}_b) = \mathbb{E}_{q_{\boldsymbol{\phi}_a}(\mathbf{s}_a,\mathbf{c}_a|\mathbf{x}_a)}\left[\log{p_{\boldsymbol{\theta}_a}(\mathbf{x}_a|\mathbf{s}_a,\mathbf{c}_a)} \right] + \mathbb{E}_{q_{\boldsymbol{\phi}_b}(\mathbf{s}_b,\mathbf{c}_b|\mathbf{x}_b)}\left[\log{p_{\boldsymbol{\theta}_b}(\mathbf{x}_b|\mathbf{s}_b,\mathbf{c}_b)} \right]  \\[.05in]
    -\mathbb{E}_{q_{\boldsymbol{\phi}_a}(\mathbf{c}_a|\mathbf{x}_a)}\left[D_{KL}\left(q_{\boldsymbol{\phi}_a}(\mathbf{s}_a|\mathbf{c}_a,\mathbf{x}_a) \| p(\mathbf{s}_a|\mathbf{c}_a) \right) \right] - \mathbb{E}_{q_{\boldsymbol{\phi}_b}(\mathbf{c}_b|\mathbf{x}_b)}\left[D_{KL}\left(q_{\boldsymbol{\phi}_b}(\mathbf{s}_b|\mathbf{c}_b,\mathbf{x}_b) \| p(\mathbf{s}_b|\mathbf{c}_b) \right) \right] \\[.05in]
    - \mathbb{E}_{q_{\boldsymbol{\phi}_a}(\mathbf{s}_a|\mathbf{c}_a,\mathbf{x}_a)}\left[D_{KL}\left(q_{\boldsymbol{\phi}_a}(\mathbf{c}_a|\mathbf{x}_a) \| p(\mathbf{c}_a) \right) \right] - \mathbb{E}_{q_{\boldsymbol{\phi}_b}(\mathbf{s}_b|\mathbf{c}_b,\mathbf{x}_b)}\left[D_{KL}\left(q_{\boldsymbol{\phi}_b}(\mathbf{c}_b|\mathbf{x}_b) \| p(\mathbf{c}_b) \right) \right] 
\end{multline}
Since all arms receive augmented samples from the same original distribution, we have $p(\mathbf{c}_a) = p(\mathbf{c}_b) = p(\mathbf{c})$. Using a simplified notation, $q_a = q_{\boldsymbol{\phi}_a}(\mathbf{c}_a|\mathbf{x}_a)$, the last two KL divergence terms can be expressed as,
\begin{equation}
    \begin{array}{ll}
    D_{KL}\left(q_a \| p(\mathbf{c}) \right)  + D_{KL}\left(q_b \| p(\mathbf{c}) \right)  &= \displaystyle{\sum_{\mathbf{c}_a}} q_a \log{\displaystyle{\frac{q_a}{p(\mathbf{c})}}} + \displaystyle{\sum_{\mathbf{c}_b}} q_b \log{\displaystyle{\frac{q_b}{p(\mathbf{c})}}} \\[.2in]
    &= \displaystyle{\sum_{\mathbf{c}_a} \sum_{\mathbf{c}_b}} q_a q_b \log{\displaystyle{\frac{q_a}{p(\mathbf{c})}}} + \displaystyle{\sum_{\mathbf{c}_a} \sum_{\mathbf{c}_b}} q_a q_b \log{\displaystyle{\frac{q_b}{p(\mathbf{c})}}} \\[.2in]
    &= \displaystyle{\sum_{\mathbf{c}_a} \sum_{\mathbf{c}_b}} q_a q_b \log{\displaystyle{\frac{q_aq_b}{p(\mathbf{c})}}} \\[.2in]
    \end{array}
\end{equation}
Now, if we marginalize $p(\mathbf{c})$ over the joint distribution $p(\mathbf{c}_a, \mathbf{c}_b)$, we can represent the categorical prior distribution as follows.
\begin{equation}
    \begin{array}{ll}
    p(\mathbf{c}) = \displaystyle{\sum_{\mathbf{c}_a,\mathbf{c}_b}} p(\mathbf{c}|\mathbf{c}_a, \mathbf{c}_b)p(\mathbf{c}_a, \mathbf{c}_b)
    \end{array}
\end{equation}
Since there is a categorical agreement condition i.e., $\mathbf{c}_a=\mathbf{c}_b$, $p(\mathbf{c})$ can be expressed as,
\begin{equation}
    p(\mathbf{c}) = \displaystyle{\sum_{\mathbf{m}}} p(\mathbf{c}|\mathbf{c}_a= \mathbf{c}_b=\mathbf{m})p(\mathbf{c}_a=\mathbf{c}_b=\mathbf{m})
\end{equation}
where
\begin{equation}
\label{eq:cond_p_joint}
    p(\mathbf{c}|\mathbf{c}_a=\mathbf{c}_b=\mathbf{m}) = \begin{cases}
    1 & \ \mathbf{m}=\mathbf{c} \\
    0 & \ \text{otherwise}
    \end{cases}
\end{equation}
Accordingly, under the $\mathbf{c}_a=\mathbf{c}_b$ constraint, we merge those KL divergence terms as follows: 
\begin{equation}
    \begin{array}{ll}
    D_{KL}\left(q_a \| p(\mathbf{c}) \right)  + D_{KL}\left(q_b \| p(\mathbf{c}) \right)  &=
    \displaystyle{\sum_{\mathbf{c}_a} \sum_{\mathbf{c}_b}} q_a q_b \log{\displaystyle{\frac{q_a q_b}{p(\mathbf{c}_a,\mathbf{c}_b)}}} \\[.2in]
    & = D_{KL}(q_a q_b \| p(\mathbf{c}_a,\mathbf{c}_b))
    \end{array}
\end{equation}
Finally, the optimization in Eq.~\ref{eq:modify_max_emc} can be expressed as
\begin{equation}
\label{eq:joint_emc}
    \begin{array}{cc}
    \max & \  \displaystyle{\sum_{a=1}^{A}} \
    (A-1)\left(\mathbb{E}_{q(\mathbf{s}_a,\mathbf{c}_a|\mathbf{x}_a)}\left[\log{p(\mathbf{x}_a|\mathbf{s}_a,\mathbf{c}_a)} \right] -
   \mathbb{E}_{q(\mathbf{c}_a|\mathbf{x}_a)}\left[D_{KL}\left(q(\mathbf{s}_a|\mathbf{c}_a,\mathbf{x}_a) \| p(\mathbf{s}_a|\mathbf{c}_a) \right) \right]\right) -\\
    & \hspace{.4in} \displaystyle{\sum_{a<b}}\  
    \mathbb{E}_{q(\mathbf{s}_a,\mathbf{s}_b|\mathbf{c}_a,\mathbf{c}_b,\mathbf{x}_a,\mathbf{x}_b)}\left[D_{KL}\left(q(\mathbf{c}_a|\mathbf{x}_a) q(\mathbf{c}_b|\mathbf{x}_b) \| p(\mathbf{c}_a,\mathbf{c}_b) \right) \right] \\[.1in]
     & \text{s.t.} \ \ \mathbf{c}_a = \mathbf{c}_b \ \ \ \forall a,b \in [1,A], a < b 
    \end{array}
\end{equation}
\section{Variational lower bound for cpl-mixVAE}
\label{sec:ELBO for Coupled mix-VAE}
In this section, using a pair of VAE arms, first we generalize the loss function for the single mix-VAE i.e., $\mathcal{L}_{\mathbf{s}}$ in Eq.~\ref{eq:single_cond_vae}, to the multi-arm case, and show that we can achieve the same objective function in Eq.~\ref{eq:joint_emc}. Then, we derive a relaxation for the equality constrained optimization.\\

\noindent
Given input data $\mathbf{x}_a$, an arm approximates two models $q(\mathbf{c}_a|\mathbf{x}_a)$ and $q(\mathbf{s}_a|\mathbf{x}_a,\mathbf{c}_a)$. If we use pairwise coupling to allow interactions between the arms, then, for a pair of VAE arms, $a$ and $b$, the variational lower bound obtained from the KL divergence in Equation~(\ref{eq:mixVAE_KLdist}) can be generalized as
\begin{eqnarray}
    \label{eq:pair_mixVAE_KL_1}
    \Delta(a,b) &\triangleq& D_{\mathrm{KL}}\left(q(\mathbf{s}_a, \mathbf{s}_b, \mathbf{c}_a, \mathbf{c}_b|\mathbf{x}_a, \mathbf{x}_b) \| p(\mathbf{s}_a, \mathbf{s}_b, \mathbf{c}_a, \mathbf{c}_b|\mathbf{x}_a, \mathbf{x}_b) \right) \nonumber\\
    &=& \displaystyle{\int_{\mathbf{s}_a}} \displaystyle{\int_{\mathbf{s}_b}} \displaystyle{\sum_{\mathbf{c}_a}} \displaystyle{\sum_{\mathbf{c}_b}} \ q(\mathbf{s}_a, \mathbf{s}_b|\mathbf{c}_a, \mathbf{c}_b, \mathbf{x}_a, \mathbf{x}_b)q(\mathbf{c}_a,\mathbf{c}_b|\mathbf{x}_a,\mathbf{x}_b) \nonumber\\ 
    &&\hspace{1.4in}\times \log{\displaystyle{\frac{q(\mathbf{s}_a,\mathbf{s}_b|\mathbf{c}_a, \mathbf{c}_b, \mathbf{x}_a,\mathbf{x}_b)q(\mathbf{c}_a,\mathbf{c}_b|\mathbf{x}_a,\mathbf{x}_b)}{\left(\displaystyle{\frac{p(\mathbf{x}_a,\mathbf{x}_b|\mathbf{s}_a,\mathbf{s}_b,\mathbf{c}_a,\mathbf{c}_b)p(\mathbf{s}_a, \mathbf{s}_b|\mathbf{c}_a, \mathbf{c}_b)p(\mathbf{c}_a,\mathbf{c}_b)}{p(\mathbf{x}_a,\mathbf{x}_b)}}\right)}}} \ d\mathbf{s}_a d\mathbf{s}_b
\end{eqnarray}
When each arm learns the continuous factor independent of other arms, we have $q(\mathbf{s}_a,\mathbf{s}_b|\mathbf{c}_a,\mathbf{c}_b,\mathbf{x}_a,\mathbf{x}_b) = q(\mathbf{s}_a|\mathbf{c}_a,\mathbf{x}_a)q(\mathbf{s}_b|\mathbf{c}_b,\mathbf{x}_b)$. Equivalently, for independent samples $\mathbf{x}_a$ and $\mathbf{x}_b$, we have $q(\mathbf{c}_a,\mathbf{c}_b|\mathbf{x}_a,\mathbf{x}_b) = q(\mathbf{c}_a|\mathbf{x}_a)q(\mathbf{c}_b|\mathbf{x}_b)$. Hence,

\begin{eqnarray}
    \label{eq:multi_mixVAE_KL_2}
    \hspace{-0.25in}\Delta(a,b)
    \hspace{-0.1in}&=&\hspace{-0.1in} \log{p(\mathbf{x}_a,\mathbf{x}_b)} + \displaystyle{\int_{\mathbf{s}_a}}\displaystyle{\int_{\mathbf{s}_b}} \displaystyle{\sum_{\mathbf{c}_a}}\displaystyle{\sum_{\mathbf{c}_b}} \ q(\mathbf{s}_a|\mathbf{c}_a,\mathbf{x}_a)q(\mathbf{s}_b|\mathbf{c}_b,\mathbf{x}_b)q(\mathbf{c}_a|\mathbf{x}_a)q(\mathbf{c}_b|\mathbf{x}_b)\log{\displaystyle{\frac{q(\mathbf{c}_a|\mathbf{x}_a)q(\mathbf{c}_b|\mathbf{x}_b)}{p(\mathbf{c}_a,\mathbf{c}_b)}}} \ d\mathbf{s}_a d\mathbf{s}_b \nonumber\\
    && + \displaystyle{\int_{\mathbf{s}_a}} \displaystyle{\sum_{\mathbf{c}_a}}  \ q(\mathbf{s}_a|\mathbf{c}_a,\mathbf{x}_a)q(\mathbf{c}_a|\mathbf{x}_a) \log{\displaystyle{\frac{q(\mathbf{s}_a|\mathbf{c}_a,\mathbf{x}_a)}{p(\mathbf{s}_a|\mathbf{c}_a)}}} \ d\mathbf{s}_a  + \displaystyle{\int_{\mathbf{s}_b}} \displaystyle{\sum_{\mathbf{c}_b}}  \ q(\mathbf{s}_b|\mathbf{c}_b,\mathbf{x}_b)q(\mathbf{c}_b|\mathbf{x}_b) \log{\displaystyle{\frac{q(\mathbf{s}_b|\mathbf{c}_b,\mathbf{x}_b)}{p(\mathbf{s}_b|\mathbf{c}_b)}}} \ d\mathbf{s}_b \nonumber\\
    &&-\displaystyle{\int_{\mathbf{s}_a}} \displaystyle{\sum_{\mathbf{c}_a}} q(\mathbf{s}_a|\mathbf{c}_a,\mathbf{x}_a)q(\mathbf{c}_a|\mathbf{x}_a)\log{p(\mathbf{x}_a|\mathbf{s}_a,\mathbf{c}_a)} \ d\mathbf{s}_a -\displaystyle{\int_{\mathbf{s}_b}} \displaystyle{\sum_{\mathbf{c}_b}} q(\mathbf{s}_b|\mathbf{c}_b,\mathbf{x}_b)q(\mathbf{c}_b|\mathbf{x}_b)\log{p(\mathbf{x}_b|\mathbf{s}_b,\mathbf{c}_b)} \ d\mathbf{s}_b \nonumber\\
   \\
   \Delta(a,b)\hspace{-0.1in}&=&\hspace{-0.1in} -\mathbb{E}_{q(\mathbf{c}_a|\mathbf{x}_a)}\left[\mathbb{E}_{q(\mathbf{s}_a|\mathbf{c}_a,\mathbf{x}_a)}\left[\log{p(\mathbf{x}_a|\mathbf{s}_a,\mathbf{c}_a)} \right] \right] - \mathbb{E}_{q(\mathbf{c}_b|\mathbf{x}_b)}\left[\mathbb{E}_{q(\mathbf{s}_b|\mathbf{c}_b,\mathbf{x}_b)}\left[\log{p(\mathbf{x}_b|\mathbf{s}_b,\mathbf{c}_b)} \right] \right] \nonumber\\ &&+\mathbb{E}_{q(\mathbf{c}_a|\mathbf{x}_a)}\left[D_{KL}\left(q(\mathbf{s}_a|\mathbf{c}_a,\mathbf{x}_a) \| p(\mathbf{s}_a|\mathbf{c}_a) \right) \right] + \mathbb{E}_{q(\mathbf{c}_b|\mathbf{x}_b)}\left[D_{KL}\left(q(\mathbf{s}_b|\mathbf{c}_b,\mathbf{x}_b) \| p(\mathbf{s}_b|\mathbf{c}_b) \right) \right] \nonumber\\
    &&+\mathbb{E}_{q(\mathbf{s}_a|\mathbf{c}_a,\mathbf{x}_a)}\left[\mathbb{E}_{q(\mathbf{s}_b|\mathbf{c}_b,\mathbf{x}_b)}\left[D_{KL}\left(q(\mathbf{c}_a|\mathbf{x}_a)q(\mathbf{c}_b|\mathbf{x}_b) \| p(\mathbf{c}_a,\mathbf{c}_b) \right) \right] \right] + \log{p(\mathbf{x}_a,\mathbf{x}_b)} 
\label{eq:coupled_mixVAE_elbo}
\end{eqnarray}
Therefore, the variational lower bound for a pair of coupled VAE arms can be expressed as,
\begin{eqnarray}
\label{eq:pair_ab}
    \mathcal{L}_{\mathrm{pair}}(a,b) &=& \mathbb{E}_{q(\mathbf{s}_a,\mathbf{c}_a|\mathbf{x}_a)}\left[\log{p(\mathbf{x}_a|\mathbf{s}_a,\mathbf{c}_a)} \right]+ \mathbb{E}_{q(\mathbf{s}_b,\mathbf{c}_b|\mathbf{x}_b)}\left[\log{p(\mathbf{x}_b|\mathbf{s}_b,\mathbf{c}_b)} \right]  \nonumber\\ 
    &&-\mathbb{E}_{q(\mathbf{c}_a|\mathbf{x}_a)}\left[D_{KL}\left(q(\mathbf{s}_a|\mathbf{c}_a,\mathbf{x}_a) \| p(\mathbf{s}_a|\mathbf{c}_a) \right) \right] - \mathbb{E}_{q(\mathbf{c}_b|\mathbf{x}_b)}\left[D_{KL}\left(q(\mathbf{s}_b|\mathbf{c}_b,\mathbf{x}_b) \| p(\mathbf{s}_b|\mathbf{c}_b) \right) \right] \nonumber\\
    && -\mathbb{E}_{q(\mathbf{s}_a|\mathbf{c}_a,\mathbf{x}_a)}\left[\mathbb{E}_{q(\mathbf{s}_b|\mathbf{c}_b,\mathbf{x}_b)}\left[D_{KL}\left(q(\mathbf{c}_a|\mathbf{x}_a)q(\mathbf{c}_b|\mathbf{x}_b) \| p(\mathbf{c}_a,\mathbf{c}_b) \right) \right] \right]
\end{eqnarray}
which is equivalent to the loss function in Eq.~\ref{eq:joint_emc}, for $A=2$.\\

\noindent
To compute the joint distribution $p(\mathbf{c}_a,\mathbf{c}_b)$, here, we define an auxiliary continuous random variable $e$ representing the mismatch (error) between $\mathbf{c}_a$ and $\mathbf{c}_b$ such that
$\forall \mathbf{c}_a,\mathbf{c}_b \in \mathcal{S}^K$, and $0<\epsilon \ll 1$,
\begin{equation}
    p(\mathbf{c}_a,\mathbf{c}_b | e) = \begin{cases}
    1 & \ |e - d^2(\mathbf{c}_a , \mathbf{c}_b)| < \epsilon/2 \\
    0 & \ \text{otherwise}
    \end{cases}
\end{equation}
Here, $d(\mathbf{c}_a, \mathbf{c}_b)$ denotes the distance between $\mathbf{c}_a$ and $\mathbf{c}_b$ in the simplex $\mathcal{S}^K$, as a measure of mismatch between categorical variables. The random variable $e$ is distributed according to an exponential probability density function with parameter $\lambda$ i.e., $\forall e \ge 0$, $f(e,\lambda)=\lambda \exp{\left(-\lambda e\right)}$, where $\lambda > 0$.
Accordingly, the joint categorical distribution can be represented as, 
%
%
\begin{eqnarray}
    p(\mathbf{c}_a,\mathbf{c}_b) &=& \displaystyle{\int} p(\mathbf{c}_a, \mathbf{c}_b | e) p(e) de \label{eq:p_joint} \\
    &=&\displaystyle{\int_{-\epsilon/2 + d^2(\mathbf{c}_a , \mathbf{c}_b)}^{\epsilon/2 + d^2(\mathbf{c}_a , \mathbf{c}_b)}} f(e, \lambda) de = \epsilon f\left(d^2\left(\mathbf{c}_a , \mathbf{c}_b\right), \lambda \right) + E \label{eq:p_error}
\end{eqnarray}
where $E$ is the error bound of the Midpoint integral rule. For given exponential function $f(e,\lambda)$, since $|f''(e,\lambda)| \le \lambda^3$, $\forall e > 0$, the Midpoint approximation error is bounded by,
\begin{equation}
    |E| \le \displaystyle{\frac{\left(\lambda \epsilon \right)^3}{24}}.
\end{equation}
Subsequently, the joint probability distribution is equivalent to:
\begin{equation}
     p(\mathbf{c}_a, \mathbf{c}_b) = \epsilon \lambda \exp{\left(-\lambda d^2\left(\mathbf{c}_a , {\mathbf{c}_b}\right)\right)} + E
\end{equation}
where $\epsilon$ and $\lambda$ are arbitrarily constant values. We can approximate the joint distribution as follows.
\begin{equation}
\label{eq:p_error_approx}
    p(\mathbf{c}_a, \mathbf{c}_b) \approx \epsilon \lambda\exp{\left(-\lambda d^2\left(\mathbf{c}_a , {\mathbf{c}_b}\right)\right)}
\end{equation}
%
%
%
Thus, the last KL divergence in Eq.~\ref{eq:pair_ab} can be approximated as,
\begin{eqnarray}
    \hspace{-0.2in}D_{\mathrm{KL}}\left(q(\mathbf{c}_a|\mathbf{x}_a)q(\mathbf{c}_a|\mathbf{x}_b) \| p(\mathbf{c}_a,\mathbf{c}_b) \right) \hspace{-0.1in}&=&\hspace{-0.1in} \displaystyle{\sum_{\mathbf{c}_a}}\displaystyle{\sum_{\mathbf{c}_b}} q(\mathbf{c}_a|\mathbf{x}_a)q(\mathbf{c}_b|\mathbf{x}_b) \log{\displaystyle{\frac{q(\mathbf{c}_a|\mathbf{x}_a)q(\mathbf{c}_b|\mathbf{x}_b)}{p(\mathbf{c}_a,\mathbf{c}_b)}}} \label{eq:KL_c}\\
    &=& -H(\mathbf{c}_a|\mathbf{x}_a) - H(\mathbf{c}_b|\mathbf{x}_b) -  \displaystyle{\sum_{\mathbf{c}_a}}\displaystyle{\sum_{\mathbf{c}_b}} q(\mathbf{c}_a|\mathbf{x}_a) q(\mathbf{c}_b|\mathbf{x}_b) \log{p(\mathbf{c}_a,\mathbf{c}_b)} \nonumber\\
    &\approx& -H(\mathbf{c}_a|\mathbf{x}_a) - H(\mathbf{c}_b|\mathbf{x}_b) + \lambda\mathbb{E}_{q(\mathbf{c}_a|\mathbf{x}_a)}\mathbb{E}_{q(\mathbf{c}_b|\mathbf{x}_b)}\left[d^2 \left(\mathbf{c}_a, \mathbf{c}_b \right)\right] - \log{\epsilon \lambda},
\label{eq:joint_KL}
\end{eqnarray}
%
%
%
Therefore, the approximated variational cost for a pair of VAE arms can be written as follows:
\begin{eqnarray}
    \mathcal{L}_{\mathrm{pair}}(a,b) &=& \mathbb{E}_{q(\mathbf{s}_a,\mathbf{c}_a|\mathbf{x}_a)}\left[\log{p(\mathbf{x}_a|\mathbf{s}_a,\mathbf{c}_a)} \right] + \mathbb{E}_{q(\mathbf{s}_b,\mathbf{c}_b|\mathbf{x}_b)}\left[\log{p(\mathbf{x}_b|\mathbf{s}_b,\mathbf{c}_b)} \right]  \nonumber\\ 
    &&-\mathbb{E}_{q(\mathbf{c}_a|\mathbf{x}_a)}\left[D_{KL}\left(q(\mathbf{s}_a|\mathbf{c}_a,\mathbf{x}_a) \| p(\mathbf{s}_a|\mathbf{c}_a) \right) \right] - \mathbb{E}_{q(\mathbf{c}_b|\mathbf{x}_b)}\left[D_{KL}\left(q(\mathbf{s}_b|\mathbf{c}_b,\mathbf{x}_b) \| p(\mathbf{s}_b|\mathbf{c}_b) \right) \right] \nonumber\\
    && +H(\mathbf{c}_a | \mathbf{x}_a) + H(\mathbf{c}_b | \mathbf{x}_b) - \lambda \mathbb{E}_{q(\mathbf{c}_a |\mathbf{x}_a)}\mathbb{E}_{q(\mathbf{c}_b |\mathbf{x}_b)}\left[d^2\left(\mathbf{c}_a, \mathbf{c}_b \right) \right]
\label{eq:aprox_coupled_mixVAE_elbo}
\end{eqnarray}
%
Now, by extending $\mathcal{L}_{pair}$ from two arms to $A$ arms, in which there are $\binom{A}{2}$ paired networks, the total loss function for $A$ arms can be written as
\begin{eqnarray}
    \mathcal{L}_{\mathrm{cpl}} &=& \sum_{a=1}^{A-1}\sum_{b=a+1}^{A} \mathcal{L}_{\mathrm{pair}}(a,b) \nonumber\\
    &=& \displaystyle{\sum_{a=1}^{A}} \
    (A-1)\mathbb{E}_{q(\mathbf{s}_a,\mathbf{c}_a|\mathbf{x}_a)}\left[\log{p(\mathbf{x}_a|\mathbf{s}_a,\mathbf{c}_a)} \right] -
   (A-1)\mathbb{E}_{q(\mathbf{c}_a|\mathbf{x}_a)}\left[D_{KL}\left(q(\mathbf{s}_a|\mathbf{c}_a,\mathbf{x}_a) \| p(\mathbf{s}_a|\mathbf{c}_a) \right) \right] \nonumber\\
    && + \displaystyle{\sum_{a < b}} \ H(\mathbf{c}_a | \mathbf{x}_a) + H(\mathbf{c}_b | \mathbf{x}_b) - \lambda \mathbb{E}_{q(\mathbf{c}_a |\mathbf{x}_a)}\mathbb{E}_{q(\mathbf{c}_b |\mathbf{x}_b)}\left[d^2\left(\mathbf{c}_a ,\mathbf{c}_b \right) \right]. \label{eq:total_coupled_mixVAE_elbo}
\end{eqnarray}
\section{Proof of Proposition 3}
\subsection{Aitchison geometry}
In this section, we first briefly review some critical definitions in \textit{Aitchison geometry}. Then, to support the proof of Proposition~\ref{prop:clr_distance_approx}, here we introduce Lemma~\ref{lem:clr_subtract} and Propositions~\ref{prop:clr_distance_approx_1} and \ref{prop:clr_distance_approx_2}.\\[.1in]

\noindent
According to {Aitchison geometry}, a simplex of $K$ parts can be considered as a vector space $(\mathcal{S}^{K}, \oplus, \otimes)$, in which $\oplus$ and $\otimes$ corresponds to \textit{perturbation} and \textit{power} operations, respectively, as follows.
\begin{equation*}
\label{eq:perturbation}
    Perturbation: \ \forall \mathbf{x}, \mathbf{y} \in \mathcal{S}^{K}, \ \mathbf{x} \oplus \mathbf{y} = \mathcal{C}\left(x_1 y_1, \dots, x_K y_K\right)
\end{equation*}
\begin{equation*}
\label{eq:power}
    Power: \ \forall \mathbf{x}\in \mathcal{S}^{K} \ and \ \forall \alpha \in \mathbb{R}, \ \alpha \otimes \mathbf{x} = \mathcal{C}\left(x_1^\alpha, \dots, x_K^\alpha\right)
\end{equation*}
where $\mathcal{C}$ denotes the closure operation as follows.
\begin{equation*}
\label{eq:closure}
   \mathcal{C}(\mathbf{x}) = \left(\displaystyle{\frac{c x_1}{\displaystyle{\sum_{k=1}^K}x_k}}, \dots, \displaystyle{\frac{c x_K}{\displaystyle{\sum_{k=1}^K}x_k}}\right).\\[.1in]
\end{equation*}
In the simplex vector space, for any $\mathbf{x}, \mathbf{y} \in \mathcal{S}^K$, the distance is defined as,
\begin{equation}
   \label{eq:Aitchison_dist}
    \begin{array}{ll}
    d_{\mathcal{S}^K}\left(\mathbf{x}, \mathbf{y}\right) &= \left(\displaystyle{\frac{1}{K}} \displaystyle{\sum_{i < j}}\left( \log{\displaystyle{\frac{x_i}{x_j}}} - \log{\displaystyle{\frac{y_i}{y_j}}}\right)^2\right)^{1/2}.
    \end{array}
\end{equation}
Furthermore, Aitchison has introduced \textit{centered-logratio} transformation (CLR), which is an isometric transformation from a simplex to a $K-$dimensional real space, $clr(\mathbf{x}) \in \mathbb{R}^{K}$. The CLR transformation involves the logratio of each $x_k$ over geometric means in the simplex as follows.
\begin{equation}
   \label{eq:clr}
    \begin{array}{ll}
    clr(\mathbf{x}) = \left(\log{\displaystyle{\frac{x_1}{g(\mathbf{x})}}}, \dots, \log{\displaystyle{\frac{x_{K}}{g(\mathbf{x})}}} \right).
    \end{array}
\end{equation}
where $g(\mathbf{x})=\left(\displaystyle{\prod_{k=1}^K}x_k\right)^{1/K}$ and $\displaystyle{\sum_{k=1}^{K}} \log{\displaystyle{\frac{x_k}{g(\mathbf{x})}}} = 0$. \\[.1in]
\noindent
Since CLR is an isometric transformation, we have
\begin{equation*}
   \label{eq:clr_isometric}
    \begin{array}{ll}
    d_{\mathcal{S}^K}\left(\mathbf{x} , \mathbf{y}\right) &= d_{\mathbb{R}^{K}}\left( clr({\mathbf{x}}) , clr({\mathbf{y}}) \right) \\[.1in]
    & = \| clr(\mathbf{x}) - clr(\mathbf{y})  \|_2 \ .
    \end{array}
\end{equation*}
The algebraic-geometric definition of $\mathcal{S}^K$ satisfies standard properties, such as
\begin{equation}
   \label{eq:dist_sum}
    \begin{array}{ll}
    d_{\mathcal{S}^K}\left(\mathbf{x} \oplus \mathbf{v}, \mathbf{y} \oplus \mathbf{v}\right) = d_{\mathcal{S}^K}\left(\mathbf{x} \ominus \mathbf{v}, \mathbf{y} \ominus \mathbf{v}\right) =  d_{\mathcal{S}^K}\left(\mathbf{x}, \mathbf{y}\right) 
    \end{array}
\end{equation}
where $\mathbf{v} \in \mathcal{S}^K$ could be any arbitrary vector in the simplex.\\[.1in]
\begin{lemma}
    \label{lem:clr_subtract}
    Given a set of vectors $\{\mathbf{x}_1, \dots, \mathbf{x}_N\}$ $\in \mathcal{S}^K$ where $\mathcal{S}^K$ is a simplex of $K$ parts, then 
    \begin{equation*}
    \centering
        clr(\mathbf{x}_1 \oplus \mathbf{x}_2 \oplus \dots \oplus \mathbf{x}_N) = clr(\mathbf{x}_1) + clr(\mathbf{x}_2) + \dots + clr(\mathbf{x}_N).
    \end{equation*}
\end{lemma}
\begin{proof}
   According to Aitchison geometry, addition of vectors in the simplex is defined as,
    \begin{equation}
    \begin{array}{ll}
        \mathbf{x}_1 \oplus \dots \oplus \mathbf{x}_N =\left(\displaystyle{\frac{\displaystyle{\prod_{n=1}^N}x_{n_1}}{c_N}}, \dots, \displaystyle{\frac{\displaystyle{\prod_{n=1}^N}x_{n_K}}{c_N}} \right)
    \end{array}
    \end{equation}
    where $c_{N} = \displaystyle{\sum_{k=1}^K \prod_{n=1}^N x_{n_k}}$.
    
    \noindent
    By applying the centered-logratio transformation, we have
    \begin{equation}
    \label{eq:clr_sum}
    \begin{array}{ll}
        clr\left(\mathbf{x}_1 \oplus \dots \oplus \mathbf{x}_N\right) =\left(\log{\displaystyle{\frac{\displaystyle{\prod_{n=1}^N}x_{n_1}}{\delta_{K,N}}}}, \dots, \displaystyle{\frac{\displaystyle{\prod_{n=1}^N}x_{n_K}}{\delta_{K,N}}} \right)
    \end{array}
    \end{equation}
    where $\delta_{K,N} = c_N \left(\displaystyle{\prod_{k=1}^K} \displaystyle{\frac{\displaystyle{\prod_{n=1}^N x_{n_k}}}{c_N}}\right)^{1/K} = \left(\displaystyle{\prod_{k=1}^K \prod_{n=1}^N x_{n_k}}\right)^{1/K}$.
    
    \noindent
    Now, we can rewrite Eq.~\ref{eq:clr_sum} as,
    \begin{equation}
    \label{eq:clr_sum_2}
    \begin{array}{ll}
        clr\left(\mathbf{x}_1 \oplus \dots \oplus \mathbf{x}_N\right) &=\left(\log{\displaystyle{\frac{x_{1_1} \dots x_{N_1}}{\left( \displaystyle{\prod_k x_{1_k}}\right)^{1/K} \dots \left( \displaystyle{\prod_k x_{N_k}}\right)^{1/K}}}}, \dots, \log{\displaystyle{\frac{x_{1_K} \dots x_{N_K}}{\left( \displaystyle{\prod_k x_{1_k}}\right)^{1/K} \dots \left( \displaystyle{\prod_k x_{N_k}}\right)^{1/K}}}} \right) \\[.5in]
        &=\left(\displaystyle{\sum_n}\log{\displaystyle{\frac{x_{n_1}}{\left(\displaystyle{\prod_k} x_{n_k}\right)^{1/K}}}}, \dots, \displaystyle{\sum_n}\log{\displaystyle{\frac{x_{n_K}}{\left(\displaystyle{\prod_k} x_{n_k}\right)^{1/K}}}} \right) \\[.5in]
        &= clr(x_1) + \dots + clr(x_N)
    \end{array}
    \end{equation}
\end{proof}
\noindent
\stepcounter{prop}
\begin{prop}
    \label{prop:clr_distance_approx_1}
    Given vectors $\mathbf{x},\mathbf{y}, \mathbf{v}_x, \mathbf{v}_y \in \mathcal{S}^K$ where $\mathcal{S}^K$ is a simplex of $K>0$ parts, then 
    \begin{equation*}
    \centering
        d^2_{\mathcal{S}^K}\left(\mathbf{x}, \mathbf{y} \right)  - \Gamma_l \leq d^2_{\mathcal{S}^K}\left(\mathbf{x} \oplus \mathbf{v}_x, \mathbf{y} \oplus \mathbf{v}_y \right) \leq d^2_{\mathcal{S}^K}\left(\mathbf{x}, \mathbf{y} \right)  + \Gamma_u
    \end{equation*}
    where $\Gamma_u, \Gamma_l \geq 0$, $\Gamma_u= K\tau_u^2 - \displaystyle{\frac{\Delta^2}{K}}$, $\Gamma_l=  \displaystyle{\frac{\Delta^2}{K}} - K\tau_l^2$, $\tau_u = \displaystyle{\max_{k}}\{\log{\displaystyle{\frac{{v}_{x_k}}{v_{y_k}}}}\}$, $\tau_l = \displaystyle{\max_{k}}\{\log{\displaystyle{\frac{v_{y_k}}{v_{x_k}}}}\}$, and $\Delta = \displaystyle{\sum_k} \left(\log{\displaystyle{\frac{v_{x_k}}{v_{y_k}}}}\right)$.
\end{prop}
\begin{proof}
   According to Aitchison geometry, the distance between two vectors $\mathbf{x}$, $\mathbf{y}$ $\in \mathcal{S}^K$ is defined as,
    \begin{equation*}
    \begin{array}{ll}
        d^2_{\mathcal{S}^K}\left(\mathbf{x}, \mathbf{y} \right) = \| clr(\mathbf{x}) - clr(\mathbf{y}) \|^2_2 
    \end{array}
    \end{equation*}
    If we perturb vectors $\mathbf{x}$ and $\mathbf{y}$ by $\mathbf{v}_x$ and $\mathbf{v}_y$, the distance between the perturbed vectors in the simple can be expressed as,
    \begin{equation*}
    \begin{array}{ll}
        d^2_{\mathcal{S}^K}\left(\mathbf{x} \oplus \mathbf{v}_x, \mathbf{y} \oplus \mathbf{v}_y \right) = \| clr(\mathbf{x} \oplus \mathbf{v}_x) - clr(\mathbf{y} \oplus \mathbf{v}_y) \|^2_2 \ .
    \end{array}
    \end{equation*}
    According to Lemma~\ref{lem:clr_subtract},
    \begin{equation}
    \begin{array}{ll}
        d^2_{\mathcal{S}^K}\left(\mathbf{x} \oplus \mathbf{v}_x, \mathbf{y} \oplus \mathbf{v}_y \right) &= \| \left(clr\left(\mathbf{x}\right) - clr\left(\mathbf{y}\right)\right) + \left(clr\left(\mathbf{v}_x\right) - clr\left(\mathbf{v}_y\right)\right) \|^2_2 \\[.1in]
        &= \| clr\left(\mathbf{x}\right) - clr\left(\mathbf{y}\right) \|_2^2  + \|clr\left(\mathbf{v}_x\right) - clr\left(\mathbf{v}_y\right)\|^2_2 +\\
        & \hspace{.1in} \left(clr\left(\mathbf{x}\right) - clr\left(\mathbf{y}\right) \right)^T \left(clr\left(\mathbf{v}_x\right) - clr\left(\mathbf{v}_y\right)\right) + \left(clr\left(\mathbf{v}_x\right) - clr\left(\mathbf{v}_y\right) \right)^T \left(clr\left(\mathbf{x}\right) -  clr\left(\mathbf{y}\right)\right) \\[.1in]
        &= d^2_{\mathcal{S}^K}\left(\mathbf{x}, \mathbf{y} \right) + d^2_{\mathcal{S}^K}\left(\mathbf{v}_x, \mathbf{v}_y \right) + 2\displaystyle{\sum_{k=1}^K}\left(\log{\displaystyle{\frac{x_k}{g(\mathbf{x})}}} - \log{\displaystyle{\frac{y_k}{g(\mathbf{y})}}} \right)\left( \log{\displaystyle{\frac{v_{x_k}}{g(\mathbf{v}_x)}}} - \log{\displaystyle{\frac{v_{y_k}}{g(\mathbf{v}_y)}}}\right)
    \end{array}
    \end{equation}
    For simplicity, let's define $d^2_1 =d^2_{\mathcal{S}^K}\left(\mathbf{x}, \mathbf{y} \right)$ and $d^2_2=d^2_{\mathcal{S}^K}\left(\mathbf{x} \oplus \mathbf{v}_x, \mathbf{y} \oplus \mathbf{v}_y \right)$, then
    \begin{equation}
    \begin{array}{ll}
        d^2_2  &= d^2_1 + d^2_{\mathcal{S}^K}\left(\mathbf{v}_x, \mathbf{v}_y \right) + 2\displaystyle{\sum_{k=1}^K}\left(\log{\displaystyle{\frac{x_k}{g(\mathbf{x})}}} - \log{\displaystyle{\frac{y_k}{g(\mathbf{y})}}} \right)\left( \log{\displaystyle{\frac{v_{x_k}}{g(\mathbf{v}_x)}}} - \log{\displaystyle{\frac{v_{y_k}}{g(\mathbf{v}_y)}}}\right) \\[.1in]
        &= d^2_1 + d^2_{\mathcal{S}^K}\left(\mathbf{v}_x, \mathbf{v}_y \right) + 2\displaystyle{\sum_{k=1}^K}\log{\displaystyle{\frac{x_k}{g(\mathbf{x})}}} \left( \log{\displaystyle{\frac{v_{x_k}}{v_{y_k}}}} - \log{\displaystyle{\frac{g(\mathbf{v}_x)}{g(\mathbf{v}_y)}}}\right) -  2\displaystyle{\sum_{k=1}^K}\log{\displaystyle{\frac{y_k}{g(\mathbf{y})}}} \left( \log{\displaystyle{\frac{v_{x_k}}{v_{y_k}}}} - \log{\displaystyle{\frac{g(\mathbf{v}_x)}{g(\mathbf{v}_y)}}}\right)
    \end{array}
    \end{equation}
    Let define $\log{\displaystyle{\frac{g(\mathbf{v}_x)}{g(\mathbf{v}_y)}}} =\log{\displaystyle{\frac{\left(\displaystyle{\prod_k}{v}_{x_k}\right)^{1/K}}{\left(\displaystyle{\prod_k}{v}_{y_k}\right)^{1/K}}}}= \displaystyle{\frac{1}{K}}\displaystyle{\sum_k}\log{\displaystyle{\frac{v_{x_k}}{v_{y_k}}}}= \displaystyle{\frac{\Delta}{K}}$. Then,
    \begin{equation}
    \begin{array}{ll}
        d^2_2 &= d^2_1 + d^2_{\mathcal{S}^K}\left(\mathbf{v}_x, \mathbf{v}_y \right) + 2\displaystyle{\sum_{k=1}^K}\log{\displaystyle{\frac{v_{x_k}}{v_{y_k}}}} \left(\log{\displaystyle{\frac{x_k}{g(\mathbf{x})}}} - \log{\displaystyle{\frac{y_k}{g(\mathbf{y})}}}\right) - \displaystyle{\frac{2\Delta}{K}}\displaystyle{\sum_{k=1}^K}\left(\log{\displaystyle{\frac{x_k}{g(\mathbf{x})}}} - \log{\displaystyle{\frac{y_k}{g(\mathbf{y})}}}\right)
    \end{array}
    \end{equation}
    Since CLR is a zero-mean transformation, $\displaystyle{\sum_k}\log{\displaystyle{\frac{x_k}{g(\mathbf{x})}}}=0$ and $\displaystyle{\sum_k}\log{\displaystyle{\frac{y_k}{g(\mathbf{y})}}}=0$. Therefore,
    \begin{equation}
    \begin{array}{ll}
        d^2_2 &= d^2_1 + d^2_{\mathcal{S}^K}\left(\mathbf{v}_x, \mathbf{v}_y \right) + 2\displaystyle{\sum_{k=1}^K}\log{\displaystyle{\frac{v_{x_k}}{v_{y_k}}}} \left(\log{\displaystyle{\frac{x_k}{g(\mathbf{x})}}} - \log{\displaystyle{\frac{y_k}{g(\mathbf{y})}}}\right) 
    \end{array}
    \end{equation}
    Additionally, $d^2_{\mathcal{S}^K}\left(\mathbf{v}_x, \mathbf{v}_y \right) \geq 0$ can be expressed as,
    \begin{equation}
    \label{eq:dist_u_upbound}
    \begin{array}{ll}
        d^2_{\mathcal{S}^K}\left(\mathbf{v}_x, \mathbf{v}_y \right)  &= \displaystyle{\sum_{k=1}^K} \left(\log{\displaystyle{\frac{v_{x_k}}{v_{y_k}}}} - \log{\displaystyle{\frac{g(\mathbf{v}_x)}{g(\mathbf{v}_y)}}}\right)^{2} \\[.1in]
        &= \displaystyle{\sum_{k=1}^K} \left(\log{\displaystyle{\frac{v_{x_k}}{v_{y_k}}}}\right)^2 + \displaystyle{\sum_{k=1}^K} \left(\log{\displaystyle{\frac{g(\mathbf{v}_x)}{g(\mathbf{v}_y)}}}\right)^{2}  - 2\log{\displaystyle{\frac{g(\mathbf{v}_x)}{g(\mathbf{v}_y)}}}\displaystyle{\sum_{k=1}^K}\log{\displaystyle{\frac{v_{x_k}}{v_{y_k}}}}\\[.1in]
        &= \displaystyle{\sum_{k=1}^K} \left(\log{\displaystyle{\frac{v_{x_k}}{v_{y_k}}}}\right)^2 - \displaystyle{\frac{\Delta^2}{K}} 
    \end{array}
    \end{equation}
    Therefore,
    \begin{equation}
    \begin{array}{ll}
        d^2_2 &= d^2_1 + \displaystyle{\sum_{k=1}^K} \left(\log{\displaystyle{\frac{v_{x_k}}{v_{y_k}}}}\right)^2 - \displaystyle{\frac{\Delta^2}{K}}  + 2\displaystyle{\sum_{k=1}^K}\log{\displaystyle{\frac{v_{x_k}}{v_{y_k}}}} \left(\log{\displaystyle{\frac{x_k}{g(\mathbf{x})}}} - \log{\displaystyle{\frac{y_k}{g(\mathbf{y})}}}\right) 
    \end{array}
    \end{equation}
    Now, consider $\tau_u = \max \{\log{\displaystyle{\frac{v_{x_k}}{v_{y_k}}}}\}$ and  $\tau_l = \max \{\log{\displaystyle{\frac{v_{y_k}}{v_{x_k}}}}\}= -\min \{\log{\displaystyle{\frac{v_{x_k}}{v_{y_k}}}}\}$. Then, 
    \begin{equation}
    \begin{array}{ll}
        d^2_2 \leq d^2_1 + K\tau^2_u - \displaystyle{\frac{\Delta^2}{K}} + 2\tau_u \left(\displaystyle{\sum_k}\log{\displaystyle{\frac{x_k}{g(\mathbf{x})}}} - \displaystyle{\sum_k}\log{\displaystyle{\frac{y_k}{g(\mathbf{y})}}}\right) \\[.15in]
         d^2_2 \geq d^2_1 + K\tau^2_l - \displaystyle{\frac{\Delta^2}{K}} - 2\tau_l \left(\displaystyle{\sum_k}\log{\displaystyle{\frac{x_k}{g(\mathbf{x})}}} - \displaystyle{\sum_k}\log{\displaystyle{\frac{y_k}{g(\mathbf{y})}}}\right)
    \end{array}
    \end{equation}
    Again, because of $\displaystyle{\sum_k}\log{\displaystyle{\frac{x_k}{g(\mathbf{x})}}}=0$ and $\displaystyle{\sum_k}\log{\displaystyle{\frac{y_k}{g(\mathbf{y})}}}=0$, we can conclude that,
    \begin{equation}
    \begin{array}{ll}
       d^2_1  - \displaystyle{\frac{\Delta^2}{K}} + K\tau_l^2 \leq d^2_2 \leq  d^2_1 - \displaystyle{\frac{\Delta^2}{K}} + K\tau_u^2 
    \end{array}
    \end{equation}
    \begin{equation}
    \begin{array}{ll}
       d^2_1  - \Gamma_l \leq d^2_2 \leq  d^2_1 +\Gamma_u
    \end{array}
    \end{equation}
Since $K\tau_u \geq \Delta \geq K\tau_l$, we can conclude $\Gamma_u, \Gamma_l \geq 0$.
\end{proof}
\noindent
\begin{prop}
    \label{prop:clr_distance_approx_2}
    Given samples $\mathbf{x},\mathbf{y} \in \mathcal{S}^K$, where $\mathcal{S}^K$ is a simplex of $K$ parts, we have 
    \begin{equation*}
    \centering
        0 \leq d^2_{\mathbf{v}}\left(\mathbf{x}, \mathbf{y} \right) - d^2_{\mathcal{S}^K}\left(\mathbf{x} \oplus \mathbf{v}_x, \mathbf{y} \oplus \mathbf{v}_y \right) \leq \displaystyle{\frac{1}{K}} (\Delta + K\tau)^2
    \end{equation*}
    where $d^2_{\mathbf{v}}\left(\mathbf{x}, \mathbf{y} \right) = \displaystyle{\sum_k}\left(\log{x_{k}v_{x_k}} - \log{y_{k}v_{y_k}} \right)^2$, $\tau = \displaystyle{\max_{k}}\{\log{\displaystyle{\frac{x_k}{y_k}}}\}$, and $\Delta = \displaystyle{\sum_k} \left(\log{\displaystyle{\frac{v_{x_k}}{v_{y_k}}}}\right)$.
\end{prop}
\begin{proof}
    \begin{equation}
    \begin{array}{ll}
        d^2_{\mathcal{S}^K}\left(\mathbf{x} \oplus \mathbf{v}_x, \mathbf{y} \oplus \mathbf{v}_y \right) & = \displaystyle{\sum_{k=1}^K}\left( \log{{{x}_{k}v_{x_k}}} - \log{{y}_{k}v_{y_k}} - \displaystyle{\frac{1}{K}} \log{\displaystyle{\prod_{k}\frac{x_{k} v_{x_k} }{y_{k} v_{y_k}}}} \right)^2\\[.3in]
    &= \displaystyle{\sum_{k=1}^K}\left( \log{x_{k} v_{x_k}} - \log{y_k v_{y_k}} - \displaystyle{\frac{1}{K}} \displaystyle{\sum_{k}}\log{\displaystyle{\frac{x_{k} v_{x_k}}{y_{k} v_{y_k}}}} \right)^2\\[.3in]
    &= \displaystyle{\sum_{k=1}^K}\left( \log{x_{k} v_{x_k}} - \log{y_k v_{y_k}} - D \right)^2
    \end{array}
    \end{equation}
    where $D = \displaystyle{\frac{1}{K}} \displaystyle{\sum_{k}}\left(\log{x_{k} v_{x_k}} - \log{y_{k} v_{y_k}}\right)$. Therefore,
    \begin{equation*}
    \begin{array}{ll}
        d^2_{\mathcal{S}^K}\left(\mathbf{x} \oplus \mathbf{v}_x, \mathbf{y} \oplus \mathbf{v}_y \right) &= \displaystyle{\sum_{k=1}^K}\left( \log{x_k v_{x_k}} - \log{y_k v_{x_k}}\right)^2 + KD^2 -2D \displaystyle{\sum_{k=1}^K}\left( \log{x_k v_{x_k}} - \log{y_k v_{y_k}}\right)\\[.2in]
        &= d^2_{\mathbf{v}}\left(\mathbf{x}, \mathbf{y} \right) - KD^2 \\[.2in]
    \end{array}
    \end{equation*}
    \begin{equation}
    \begin{array}{ll}
        d^2_{\mathbf{v}}\left(\mathbf{x}, \mathbf{y} \right) = d^2_{\mathcal{S}^K}\left(\mathbf{x} \oplus \mathbf{v}_x, \mathbf{y} \oplus \mathbf{v}_y \right) + KD^2 \\[.2in]
    \end{array}
    \end{equation}
    Since $KD^2 \geq 0$, $d^2_{\mathbf{v}}\left(\mathbf{x}, \mathbf{y} \right) \geq d^2_{\mathcal{S}^K}\left(\mathbf{x} \oplus \mathbf{v}_x, \mathbf{y} \oplus \mathbf{v}_y \right)$.\\[.1in]
    
    \noindent
    Now, considering $\tau = \displaystyle{\max_{k}}\{\log{\displaystyle{\frac{x_k}{y_k}}}\}$, and $\Delta = \displaystyle{\sum_k} \left(\log{\displaystyle{\frac{v_{x_k}}{v_{y_k}}}}\right)$, then
    \begin{equation}
    \begin{array}{ll}
        d^2_{\mathbf{v}}\left(\mathbf{x}, \mathbf{y} \right) - d^2_{\mathcal{S}^K}\left(\mathbf{x} \oplus \mathbf{v}_x, \mathbf{y} \oplus \mathbf{v}_y \right) &= \displaystyle{\frac{1}{K}}\left(\displaystyle{\sum_k}\left(\log{\displaystyle{\frac{x_{k}}{y_{k}}}} + \log{\displaystyle{\frac{v_{x_k}}{v_{y_k}}}}\right)\right)^2 \\[.2in]
        &= \displaystyle{\frac{1}{K}}\left(\Delta + \displaystyle{\sum_k}\left(\log{\displaystyle{\frac{x_{k}}{y_{k}}}} \right)\right)^2 \\[.2in]
        &\leq \displaystyle{\frac{1}{K}}(\Delta + K\tau)^2 \ .
    \end{array}
    \end{equation}
\end{proof}

\noindent
\setcounter{prop}{2}
\begin{prop}
    \label{prop:clr_distance_approx}
     Suppose $\mathbf{c}_{a},\mathbf{c}_{b} \in \mathcal{S}^K$, where $\mathcal{S}^K$ is a simplex of $K > 0$ parts. If  $d_{S^K}\left(\mathbf{c}_{a}, \mathbf{c}_{b} \right)$ denotes the distance in Aitchison geometry and $d^2_{\sigma}(\mathbf{c}_{a},\mathbf{c}_{b}) = \sum_{k}\left(\sigma^{-1}_{a_{k}}\log{c_{a_k}} -  \sigma^{-1}_{b_k}\log{c_{b_k}}\right)^2$ denotes a perturbed distance, then
    \begin{equation*}
    \centering
        d^2_{S^K}\left(\mathbf{c}_{a}, \mathbf{c}_{b} \right) - \rho_l \leq d^2_{\sigma}\left(\mathbf{c}_{a}, \mathbf{c}_{b} \right) \leq  d^2_{S^K}\left(\mathbf{c}_{a}, \mathbf{c}_{b} \right) + \rho_u
    \end{equation*}
    where $\rho_u, \rho_l \geq 0$, $\rho_u=K\left(\tau^2_{{\sigma}_u} + \tau^2_{\mathbf{c}} \right) + 2\Delta_{{\sigma}}\tau_{\mathbf{c}}$, $\rho_l= \displaystyle{\frac{\Delta_{\sigma}^2}{K}} - K\tau^2_{{\sigma}_l}$, $\tau_{\mathbf{c}} = \displaystyle{\max_{k}}\{\log{c_{a_k}} - \log{c_{b_k}}\}$, $\tau_{{\sigma}_u}=\displaystyle{\max_{k}}\{g_k\}$, $\tau_{{\sigma}_l}=\displaystyle{\max_{k}}\{-g_k\}$, $\Delta_{\sigma}= \displaystyle{\sum_k}g_k$, and $g_k=(\sigma^{-1}_{a_k}-1)\log{c_{a_k}} - (\sigma^{-1}_{b_k}-1)\log{c_{b_k}}$.
\end{prop}
\begin{proof}
   In Propositions~\ref{prop:clr_distance_approx_1} and \ref{prop:clr_distance_approx_2}, by considering $\mathbf{x}=\mathbf{c}_{a}$, $\mathbf{y}=\mathbf{c}_{b}$, $\mathbf{v}_x=\mathbf{v}_a=\left(\displaystyle{\frac{c_{a_1}^{(\sigma^{-1}_{a_1}-1)}}{\gamma_a}}, \dots, \displaystyle{\frac{c_{a_K}^{(\sigma^{-1}_{a_K}-1)}}{\gamma_a}}\right)$, and $\mathbf{v}_y=\mathbf{v}_b=\left(\displaystyle{\frac{c_{b_1}^{(\sigma^{-1}_{b_1}-1)}}{\gamma_b}}, \dots, \displaystyle{\frac{c_{b_K}^{(\sigma^{-1}_{b_K}-1)}}{\gamma_b}}\right)$, where $\gamma_a = \displaystyle{\sum_k} c_{a_k}^{(\sigma^{-1}_{a_k}-1)}$ and $\gamma_b = \displaystyle{\sum_k} c_{b_k}^{(\sigma^{-1}_{b_k}-1)}$, we have
    \begin{equation}
    \begin{array}{ll}
       d^2_{S^K}\left(\mathbf{c}_{a} \oplus \mathbf{v}_a, \mathbf{c}_{b} \oplus \mathbf{v}_b \right) &= \displaystyle{\sum_{k=1}^K}\left( \sigma_{a_k}^{-1}\log{c_{a_k}} -  \sigma_{b_k}^{-1} \log{c_{b_k}} - D \right)^2
    \end{array}
    \end{equation}
    where $D = \displaystyle{\frac{1}{K}} \displaystyle{\sum_{k}}\left(\sigma_{a_k}^{-1}\log{c_{a_k}} -  \sigma_{b_k}^{-1} \log{c_{b_k}}\right)$. Hence,
    \begin{equation}
    \begin{array}{ll}
        d^2_{S^K}\left(\mathbf{c}_{a}, \mathbf{c}_{b} \right) + K\tau^2_{{\sigma}_l} - \displaystyle{\frac{\Delta^2_{\sigma}}{K}} \leq d^2_{S^K}\left(\mathbf{c}_{a} \oplus \mathbf{v}_a, \mathbf{c}_{b} \oplus \mathbf{v}_b \right) 
        \leq d^2_{S^K}\left(\mathbf{c}_{a}, \mathbf{c}_{b} \right) + K\tau^2_{{\sigma}_u} - \displaystyle{\frac{\Delta^2_{\sigma}}{K}}
    \end{array}
    \end{equation}
    and
    \begin{equation}
    \begin{array}{ll}
        0 \leq d^2_{\sigma}\left(\mathbf{c}_{a}, \mathbf{c}_{b} \right) - d^2_{S^K}\left(\mathbf{c}_{a} \oplus \mathbf{v}_a, \mathbf{c}_{b} \oplus \mathbf{v}_b \right)
        \leq \displaystyle{\frac{1}{K}}(K\tau_{\mathbf{c}} + \Delta_{\sigma})^2
    \end{array}
    \end{equation}
    Therefore,
    \begin{equation}
    \begin{array}{ll}
        d^2_{S^K}\left(\mathbf{c}_{a}, \mathbf{c}_{b} \right) + K\tau^2_{{\sigma}_l} - \displaystyle{\frac{\Delta^2_{\sigma}}{K}} \leq d^2_{\sigma}\left(\mathbf{c}_{a}, \mathbf{c}_{b} \right) 
        \leq d^2_{S^K}\left(\mathbf{c}_{a}, \mathbf{c}_{b} \right) + \displaystyle{\frac{1}{K}}\left(\left(K\tau_{\mathbf{c}} + \Delta_{\sigma}\right)^2 + K^2\tau^2_{{\sigma}_u} - \Delta_{\sigma}^2\right)
    \end{array}
    \end{equation}
    \begin{equation}
    \begin{array}{ll}
        d^2_{S^K}\left(\mathbf{c}_{a}, \mathbf{c}_{b} \right) - \rho_l \leq d^2_{\sigma}\left(\mathbf{c}_{a}, \mathbf{c}_{b} \right) 
        \leq d^2_{S^K}\left(\mathbf{c}_{a}, \mathbf{c}_{b} \right) + \rho_u \ .
    \end{array}
    \end{equation}
\end{proof}
\section{Type-preserving data augmentation}
\label{sec:Type-preserving augmentation}
Augmentation can be considered as a generative process. We seek a generative model that not only learns the data distribution, but also transformations that represent within-class variations in an unsupervised manner. Learning such transformations is generally not straightforward, and requires prior knowledge about the underlying invariances. While conventional transformations such as rotation, scaling, or translation can serve as type-preserving augmentations for many image datasets, they may not capture the richness of the underlying process. Moreover, such augmentation strategies cannot be used when within-class invariance are unknown. Suggested alternatives to conventional augmentations either rely on class label, or are specific to image data\footnote{Søren Hauberg, Oren Freifeld, Anders Boesen Lindbo Larsen, John Fisher, and Lars Hansen. Dreaming more data: Class-dependent distributions over diffeomorphisms for learned data augmentation.
In Artificial Intelligence and Statistics, pp. 342–350, 2016.}~\footnote{Ayush Jaiswal, Rex Yue Wu, Wael Abd-Almageed, and Prem Natarajan. Unsupervised adversarial
invariance. In Advances in Neural Information Processing Systems, pp. 5092–5102, 2018.}.\\

\noindent
To this end, inspired by DAGAN\footnote{Antoniou, Antreas, Amos Storkey, and Harrison Edwards. "Data augmentation generative adversarial networks." arXiv preprint arXiv:1711.04340, 2017.}, we propose an unsupervised type-preserving augmentation using a VAE-GAN-like architecture\footnote{Anders Boesen Lindbo Larsen, Søren Kaae Sønderby, Hugo Larochelle, and Ole Winther. Autoencoding beyond pixels using a learned similarity metric. In International conference on machine
learning, pp. 1558–1566. PMLR, 2016.}. We seek a network $\mathcal{G}$ such that a noisy copy, $\mathbf{x_a}$ can be obtained as a variation of the given sample, $\mathbf{x}$, based on its low dimensional representation that is concatenated with Gaussian noise $\mathbf{n}$. To prevent the network from disregarding the noise, we formulate the training procedure as the following minmax optimization which uses a discriminator network $\mathcal{D}$ as a regularizer.
\begin{equation}
\label{eq:udagan}
\begin{array}{cc}
     \displaystyle{\min_{\mathcal{G}}}\ \displaystyle{\max_{\mathcal{D}}} \ 
     \mathcal{V}\left(\mathcal{D}, \mathcal{G}\right) -\mathcal{R}_{\mathbf{}}(\mathcal{G}) 
    + \mathcal{T}_{\alpha}(\mathcal{G}) + \gamma d(\mathcal{G}) 
\end{array}
\end{equation}
where,
\begin{eqnarray}
    \mathcal{V}\left(\mathcal{D}, \mathcal{G} \right) &=& \mathbb{E}_{\mathbf{x}}\left[\log{\mathcal{D}(\mathbf{x})}\right] +  \mathbb{E}_{\mathbf{x}}\left[\log{(1 - \mathcal{D}(\mathbf{x}_{\not \mathbf{n}}))}\right] + \mathbb{E}_{\mathbf{x}, \mathbf{n}}\left[\log{(1 - \mathcal{D}(\mathbf{x}_{\mathbf{n}}))}\right]\\
    \mathcal{R}\left(\mathcal{G} \right) &=& \mathbb{E}_{q(\mathbf{z}|\mathbf{x})}\left[\log{p(\mathbf{x} | \mathbf{z})}\right] \\
    \mathcal{T}_{\alpha}(\mathcal{G}) &=& \max{ \left(\| \mathbf{x}- \mathbf{x}_{\not \mathbf{n}} \|_2 - \|\mathbf{x}- \mathbf{x}_{\mathbf{n}} \|_2 + \alpha, 0\right)} \\
    d\left(\mathcal{G} \right) &=& D_{KL}\left(q(\mathbf{z}|\mathbf{x}) \| q(\mathbf{z}|\mathbf{x}, \mathbf{n}) \right).
\end{eqnarray}
While training, $\mathcal{G}$ generates two samples: $\mathbf{x}_{\mathbf{n}}$ and $\mathbf{x}_{\not \mathbf{n}}$. The former denotes $\mathbf{x_a}$, and the latter is a sample generated in the absence of noise. In Eq.~\ref{eq:udagan}, $\mathcal{V}$ is the value function for the joint training of the discriminator and generator;
$\mathcal{R}$ is the reconstruction loss, which operates only over $\hat{\mathbf{x}}$; $\mathcal{T}_{\alpha}(\mathcal{G})$ is the triplet loss that prevents network $\mathcal{G}$ from disregarding noise and generating identical samples; and $d\left(\mathcal{G} \right)$ is the distance between the latent variables in the absence and presence of noise. $d\left(\mathcal{G} \right)$ is a regularizer to encourage original and noisy samples to be located close to one another in the latent space and is controlled by hyperparameter $\gamma \ll 1$. Fig.~\ref{fig:augmenterDesing} illustrates the network design for the type-preserving data augmentation for image datasets. For the scRNA-seq dataset, we used the similar design that is used for a single arm in cpl-mixVAE~(Fig.~\ref{fig:cplMixVAEDesign}b), without mixture representation, only a continuous space, with $|\mathbf{z}|=10$.
\augmenterDesing
\mnistAugmentation
\subsection{Data augmentation for an image dataset: MNIST}
Fig.~\ref{fig:mnistAugmentation} displays example noisy samples generated by the type-preserving augmentation for MNIST. To quantitatively evaluate the proposed type-preserving data augmentation, we used a benchmark classifier for MNIST digits, which achieves {$99.54\%$} accuracy over 10,000 test samples\footnote{Digit Recognizer, kaggle competition: https://www.kaggle.com/c/digit-recognizer}. Applying the imported classifier to the augmented test samples yields {$96.14\%$} classification accuracy, which demonstrates that the augmenter preserves the label information (type) for $\mathbf{96.58\%}$ of the augmented samples. 
\subsection{Data augmentation for a non-image dataset: scRNA-seq}
Generating augmented samples with the same class identity in the absence of within-class invariance is fairly challenging. In case of image datasets, e.g. MNIST, since there exist some intuitions about the identities of discrete and continuous variational factors, we can explicitly define a set of transformation such as rotation, translation, scaling, flipping, etc. that can be used as type-preserving augmentation. However, for non-image datasets, e.g. the single cell RNA-seq dataset, suggested alternative methods may fail to represent the class-conditioned variation in an unsupervised manner. Moreover, in case of biological datasets, learning an augmentation transformation is rather challenging due to the limited number of samples. Accordingly, in this section, we study the performance of the proposed data augmentation to investigate the extent to which our method is successful in realistic generation of the single-cell RNA-seq samples.\\

\snRANseqLow
\sst
\noindent
Fig.~\ref{fig:snRANseqLow} illustrates a two-dimensional demonstrations for both original and augmented single cells samples. For two-dimensional visualizations, here, we used a regular autoencoder for non-linear dimension reduction. First, the autoencoder has been trained on the original cell samples. After learning a two-dimensional coordinate system for the original samples (left panel), we used the autoencoder to visualize the augmented samples (right panel). Comparing the visualizations demonstrates that the representations are qualitatively similar and all groups of cells sharing the same type (same color) are placed in similar locations. Additionally, in Fig.~\ref{fig:sst}, we show the expression profiles of a subset of genes for an inhibitory cell. Again the qualitative comparison of the expression profiles reveals a similar variability across genes. Since the single cell RNA-seq data is heavily unbalanced, we additionally reported the data augmenter's performance at the single gene expression level. Fig.~\ref{fig:geneDistribution} illustrates the expression distribution of a subset of known genes for augmented cell samples (colorful histograms) compared with the original expressions (gray histograms). 
%
\geneDistribution
\section{MNIST dataset analysis}
\label{sec:MNIST dataset analysis}
A common assumption in ``disentangling'' the continuous and discrete factors of variability is the independence of the categorical and continuous latent variables, conditioned on data. Fig.~\ref{fig:mnistStyleHistogramSupp} demonstrates that this assumption can be significantly violated for two commonly used, interpretable style variables, ``angle'' and ``width,'' in the MNIST dataset.\\[.01in]

\noindent
{\bf Calculation of angle and width:} We first calculate the inertia matrix for each sample by treating the image as a solid object with a mass distribution given by pixel brightness values. Then, we compute the principal axis of the image based on the inertia matrix. We report the angle between this vector and the vertical axis using the $[-\pi/2, \pi/2)$ range. To calculate the width, we project the image to the horizontal axis after aligning the principal axis with the vertical axis using the computed angle value. We report the support of this projected signal, normalized by the horizontal size of the image (here 28 pixels).
\section{Dependence of state and class label in JointVAE}
We analyzed the effects of the dependency between the continuous and discrete latent factors on the results obtained by state-of-the-art methods for joint representation learning, e.g. JointVAE or CascadeVAE. These methods formulate the continuous and discrete variability as two independent factors such that the discrete factor is expected to determine the cluster to which each sample belongs, while the continuous factor represents the \textit{class-independent} variability. In many applications, however, the assumption of a discrete-continuous dichotomy may not be satisfied. (Section~\ref{sec:MNIST dataset analysis} analyzes this assumption for the MNIST dataset.) \\

\noindent
Fig.~\ref{fig:latentSpaceJointVAE}a illustrates four dimensions of the continuous latent variable $\mathbf{s}$ obtained by the JointVAE model for the MNIST dataset. Here, colors represent the digit type of each $\mathbf{s}$ sample. While the prior distribution is assumed to be Gaussian, the dependency of $\mathbf{s}|\mathbf{x}$ on the digit type, $\mathbf{c}$, is visible. To quantify this observation, we applied an unsupervised clustering method, i.e. Gaussian mixture model (GMM) with 10 clusters, to the continuous RV samples obtained from a JointVAE network trained for 150000 iterations. This unsupervised model achieved an overall clustering accuracy of $\mathbf{66\%}$. Fig.~\ref{fig:latentSpaceJointVAE}b shows the results for individual digits, e.g. $83\%$ for digit ``1'' (Fig.~\ref{fig:latentSpaceJointVAE}). Together, these results demonstrate the violation of the independence assumption for $q(\mathbf{s}|\mathbf{x})$ and $q(\mathbf{c}|\mathbf{x})$.
%
\mnistStyleHistogramSupp
%
\latentSpaceJointVAE
\section{Ablation studies}
\subsection{MNIST}
As discussed earlier in Section~4.3, to show how the $A$-arm VAE framework is successful in mixture modeling, we investigated the categorical assignment performance under different training settings. Since CascadeVAE does not learn the categorical factors by variational inference, here we only studied JointVAE as a 1-arm VAE, and cpl-mixVAE as a 2-arm VAE. Table~\ref{tab:mnistResults} shows the performance of JointVAE, cpl-mixVAE and their variants under different training settings for the MNIST dataset. Here, we considered the reconstruction error, i.e. $\mathcal{L}_{rec}$ and accuracy of the categorical performance (ACC).  
In Table~\ref{tab:mnistResults}, JointVAE denotes the average performance for the original JointVAE that has been trained by settings suggested in~(Dupont, 2018); JointVAE$^\dagger$, is the JointVAE model that has been trained with noisy copies of the original MNIST dataset generated by the type-preserving data augmentation method in Section F; JointVAE$^\ddagger$ is another JointVAE model that uses the same architecture for the basic encoder/decoder networks as the one used in cpl-mixVAE. Our results does not show any improvement in the performance of JointVAE by using data augmentation or altering the network architecture. \\

\noindent
Next, we studied the performance changes of the proposed 2-arm cpl-mixVAE under three different settings. In Table~\ref{tab:mnistResults}, cpl-mixVAE is the proposed 2-arm VAE framework using coupled-autoencoders and the type-preserving data augmentation in Section F; cpl-mixVAE$^*$, is a cpl-mixVAE in which coupled networks are not independent and the networks parameters are shared; cpl-mixVAE$^a$, is a cpl-mixVAE model that uses random rotations as an affine transformation for data augmentation; and lastly cpl-mixVAE$(\mathbf{s} \centernot\mid \mathbf{c})$, is a cpl-mixVAE model in which the state variable is independent of the discrete variable. Our results show that the proposed cpl-mixVAE obtained the best categorical assignment among all training settings. 
\mnistResults
\subsection{scRNA-seq}
Here, we examine the accuracy of categorical assignments for VAE-based mixture models, under different dimensions of discrete space. For this purpose, we merged neuron types in the scRNA-seq dataset using hierarchical taxonomy defined by~Tasic et al., 2018. Fig.~\ref{fig:taxonomy} illustrates the cell type taxonomy for the scRNA-seq dataset. The dendrogram shows the hierarchical relationship between 115 neuron types, where the first bifurcation from the top represents the split between inhibitory (right branch) and excitatory neurons (left branch). 
We used the hierarchical dendrogram to assess the performance of the 1-arm and 2-arm VAEs at different levels of cell type taxonomy. First, we obtained a smaller number of cell classes by progressively merging the 115 types according to the hierarchical dendrogram. For instance, at the three merging levels from bottom to top, indicated in Fig.~\ref{fig:taxonomy}, we obtain 57, 10, and 2 distinct merged neuron sub-classes, respectively. \\

\noindent
Fig.~\ref{fig:multiKFACS}b compares the performances of JointVAE and cpl-mixVAE(2-arm) for different numbers of categories obtained for merged types. For instance, $|\mathbf{c}|=2$ corresponds to the highest node in the dendrogram, where there are only two categories. As expected, the accuracy significantly increases as types are merged according to the hierarchy. Consistent with the results for the MNIST dataset, once again cpl-mixVAE outperforms the JointVAE model. Note that here, chance level is estimated based on the most abundant type in the dataset.\\

\noindent
Additionally, here, we report the clustering performances of both VAE arms in Fig.~\ref{fig:multiKFACS}b. Our results demonstrate that the performances of both arms are very similar, suggesting that they identify similar types with comparable accuracy. 
\multiKFACS

\section{Implementation and training settings}
\subsection{Architecture of the networks}
Fig.~\ref{fig:cplMixVAEDesign} shows the network architecture for the 2-coupled mixVAE model applied on the benchmark datasets, e.g. MNIST~(Fig.~\ref{fig:cplMixVAEDesign}a), and the scRNA-seq dataset~(Fig.~\ref{fig:cplMixVAEDesign}b), respectively. In this architecture, each VAE arm received non-identical copies of the original sample. \\

\noindent
For all dataset, To enhance the training process, we also applied $20\%$ and $10\%$ random dropout of the input sample and the state variable, respectively. \\

\noindent
For MNIST and dSprites, JointVAE and CascadeVAE have been trained by using the same network design and training parameters suggested in~(Dupont, 2018; Jeong $\&$ Song, 2019). \\

\noindent
For InfoGAN, we also used the same network design and parameter setting suggested in~(Chen et al., 2016) for the MNIST dataset. \\

\noindent
JointVAE$^\ddagger$ uses the same network architecture as a single arm of cpl-mixVAE. That is, it still uses the same loss function and learning procedure as JointVAE, but its convolutional layers are replaced by fully-connected layers, to demonstrate that these architecture choices do not explain the improvement achieved by cpl-mixVAE. \\

\noindent
Following, we provide the details of the training parameters for introduced models, for each dataset. 
Note that to calculate the computational cost of each method~(Table~1), we unified the batch size and size of data (e.g. image size) across all methods, and reported the execution time of each iteration on the same GPU machine.  
\subsection{Training parameters for the MNIST dataset}
Training details used for the MNIST dataset are listed as follows. For JointVAE$^\dagger$ and JointVAE$^\ddagger$ model, we used the same training parameters as reported in~(Dupont, 2018). \\[.1in]
\textbf{cpl-mixVAE}
\begin{itemize}
    \item Continuous and categorical variational factors: $|\mathbf{s}|=10$, $|\mathbf{c}|=10$
    \item Batch size: $256$
    \item Training epochs: $600$
    \item $\tau$ (temperature for sampling from Gumbel-softmax distribution): $0.67$
    \item $\lambda$~(coupling weight): $1$
    \item Optimizer: Adam with learning rate 1e-4
\end{itemize}
\textbf{JointVAE$^\dagger$}, \textbf{JointVAE$^\ddagger$}
\begin{itemize}
    \item Continuous and categorical variational factors: $|\mathbf{s}|=10$, $|\mathbf{c}|=10$
    \item Batch size: $64$
    \item Training epochs: $160$
    \item $\tau$: $0.67$
    \item $\gamma_\mathbf{s}$, $\gamma_\mathbf{c}$ (hyperparameters of KL divergence): 30
    \item $C_\mathbf{s} \in \mathbb{R}^{10}$, $C_\mathbf{c} \in \mathbb{R}^{10}$ (channel capacities): Increased linearly from 0 to 5 in 25000 iterations
    \item Optimizer: Adam with learning rate 1e-4
\end{itemize}
%
\cplMixVAEDesign
\subsection{Training parameters for the dSprites dataset}
Training details used for the dSprites dataset are listed as follows. \\[.05in]
\newline
\textbf{cpl-mixVAE}
\begin{itemize}
    \item Continuous and categorical variational factors: $|\mathbf{s}|=6$, $|\mathbf{c}|=3$
    \item Batch size: $256$
    \item Training epochs: $600$
    \item $\tau$: $0.67$
    \item $\lambda$~(coupling weight): $1$
    \item Optimizer: Adam with learning rate 1e-4
\end{itemize}
\subsection{Training parameters for the scRNA-seq dataset}
Training details used for the scRNA-seq dataset are listed as follows. For the JointVAE and CascadeVAE models, we used the same network architecture that we employed for each arm in cpl-mixVAE~(Fig.~\ref{fig:cplMixVAEDesign}b). We tried to set the training parameters according to the reported numbers in~(Dupont, 2018; Jeong $\&$ Song, 2019).\\[.05in]
\newline
\textbf{cpl-mixVAE}
\begin{itemize}
    \item Continuous and categorical variational factors: $|\mathbf{s}|=2$, $|\mathbf{c}|=115$
    \item Batch size: $1000$
    \item D (size of the last hidden layer): $10$
    \item Training epochs: $10000$
    \item $\tau$: $1$
    \item $\lambda$~(coupling weight): $1$
    \item Optimizer: Adam with learning rate 1e-3
\end{itemize}
\textbf{JointVAE} 
\begin{itemize}
    \item Continuous and categorical variational factors: $|\mathbf{s}|=2$, $|\mathbf{c}|=115$
    \item Batch size: $1000$
    \item D (size of the last hidden layer): $10$
    \item Training epochs: $10000$
    \item $\tau$: $1$
    \item $\gamma_\mathbf{s}$, $\gamma_\mathbf{c}$ (hyperparameters of KL divergence): 100
    \item $C_\mathbf{s}\in \mathbb{R}^2$, $C_\mathbf{c}\in \mathbb{R}^{115}$ (channel capacities): Increased linearly from 0 to 10 in 100000 iterations
    \item Optimizer: Adam with learning rate 1e-3
\end{itemize}
\textbf{CascadeVAE} 
\begin{itemize}
    \item Continuous and categorical variational factors: $|\mathbf{s}|=2$, $|\mathbf{c}|=115$
    \item Batch size: $1000$
    \item D (size of the last hidden layer): $10$
    \item Training epochs: $45000$
    \item $\lambda^{\prime}$ (hyperparameter of $D_{KL}\left(q(\mathbf{c}|\mathbf{x}) \ \| \ U(|\mathbf{c}|) \right) $): 0.1
    \item $\beta$ (hyperparameter of $D_{KL}\left(q(\mathbf{s}|\mathbf{x}) \ \| \ p(\mathbf{s}) \right) $): Increased linearly from 0 to 10 in 100000 iterations
    \item Optimizer: Adam with learning rate 1e-4
\end{itemize}